\pdfoutput=1

\documentclass[11pt]{article}

\usepackage[final]{acl}

\usepackage{times}
\usepackage{latexsym}

\usepackage[T1]{fontenc}

\usepackage[utf8]{inputenc}

\usepackage{microtype}
\usepackage{graphicx}
\usepackage{caption}
\usepackage{subcaption}
\usepackage{amsmath}
\usepackage{cleveref}
\usepackage{amssymb}
\usepackage[normalem]{ulem}
\usepackage{float}
\usepackage{multirow}
\usepackage{natbib}
\usepackage{lineno}
\usepackage[ruled,linesnumbered,vlined]{algorithm2e}
\usepackage{mathtools}
\DeclarePairedDelimiter\ceil{\lceil}{\rceil}

\usepackage{inconsolata}

\newcommand{\AB}[1]{\textcolor{blue}{\small[AB: #1]}}
\newcommand{\TV}[1]{\textcolor{red}{\small[TV: #1]}}
\newcommand{\lyc}[1]{\textcolor{orange}{\small[YL: #1]}}

\newcommand{\EXTENSION}[1]{}

\title{NeLLCom-X: A Comprehensive Neural-Agent Framework \\ to Simulate Language Learning and Group Communication}

\author{
  Yuchen Lian$^\diamond$ $^\dagger$
  \qquad
  Tessa Verhoef$^\dagger$$^{\ast}$
  \qquad
  Arianna Bisazza$^\ddagger$\thanks{Shared senior authorship.}
  \\
  \ \\
  $^\diamond$Faculty of Electronic and Information Engineering, Xi'an Jiaotong University
  \\
  $^\dagger$Leiden Institute of Advanced Computer Science, Leiden University
  \\
  \texttt{\{y.lian, t.verhoef\}@liacs.leidenuniv.nl}
  \\
  $^\ddagger$Center for Language and Cognition, University of Groningen \\
  \texttt{a.bisazza@rug.nl}
}

\date{}

\begin{document}
\maketitle
\begin{abstract}
Recent advances in computational linguistics include simulating the emergence of human-like languages with interacting neural network agents, starting from sets of random symbols. 
The recently introduced NeLLCom framework \cite{lian-etal-2023-communication} allows agents to first learn an artificial language and then use it to communicate, with the aim of studying the emergence of specific linguistics properties. 
We extend this framework (NeLLCom-X) by introducing more realistic role-alternating agents and group communication in order to investigate the interplay between language learnability, communication pressures, and group size effects. 
We validate NeLLCom-X by replicating key findings from prior research simulating the emergence of a word-order/case-marking trade-off. Next, we investigate how interaction 
affects linguistic convergence and emergence of the trade-off. The novel framework facilitates future simulations of diverse linguistic aspects, emphasizing the importance of interaction and group dynamics in language evolution.
\end{abstract}

\EXTENSION{Make all plots slightly larger}

\section{Introduction}

Human language can be viewed as a complex adaptive dynamical system \cite{fitch2007invisible, steels2000language, beckner2009language}, in which individual behaviours of language users drive linguistic emergence and change at the population level. Languages are shaped by the brains of individuals who are learning them  \cite{christiansen2008language, kirby2014iterated} and novel conventions and meanings are negotiated during interaction and language use \cite{fusaroli2012carving, namboodiripad2016measuring, garrod2007foundations}. The effect of these mechanisms on linguistic patterns has been studied extensively, and it is recognized that language systems do not spring from the mind of a single individual, but are the result of constant reinterpretation and filtering through populations of human minds. As such, language users are not mere passive learners, but unconsciously and gradually contribute to language change.

Recently, this interactive and dynamic property of human language was recognized as a key factor to improve AI \cite{mikolov2018roadmap}, leading to a large interest in simulating the emergence of human-like languages with neural network agents \cite{havrylov2017emergence, kottur2017natural, lazaridou2017multiagent, lazaridou2020emergent}. Typically, a pair of agents is simulated where a speaking agent tries to help a listener recover an intended meaning by generating a message the listener can interpret. Early frameworks have been progressively expanded to display important aspects of human language and communication, like generational transmission \cite{li-2019-easeteaching, chaabouni-etal-2019-word, lian2021effect, chaabouni2022emergent}, group interaction \cite{tieleman2019shaping, chaabouni2022emergent,rita2022emergent, michel2023revisiting, kim2021emergent} and other aspects \cite{galke2024emergent}.
Within this body of work, most studies  start from \textit{sets of random symbols}, with a strong focus on tracking the emergence of human-like language properties such as compositionality \cite{chaabouni2020compositionality,chaabouni2022emergent, li-2019-easeteaching, conklin2022compositionality} or principles of lexical organization like Zipf’s law of abbreviation \cite{rita2020lazimpa}.

\begin{figure*}[t]
    \centering
    \captionsetup{singlelinecheck=off}
  \includegraphics[width=\textwidth]{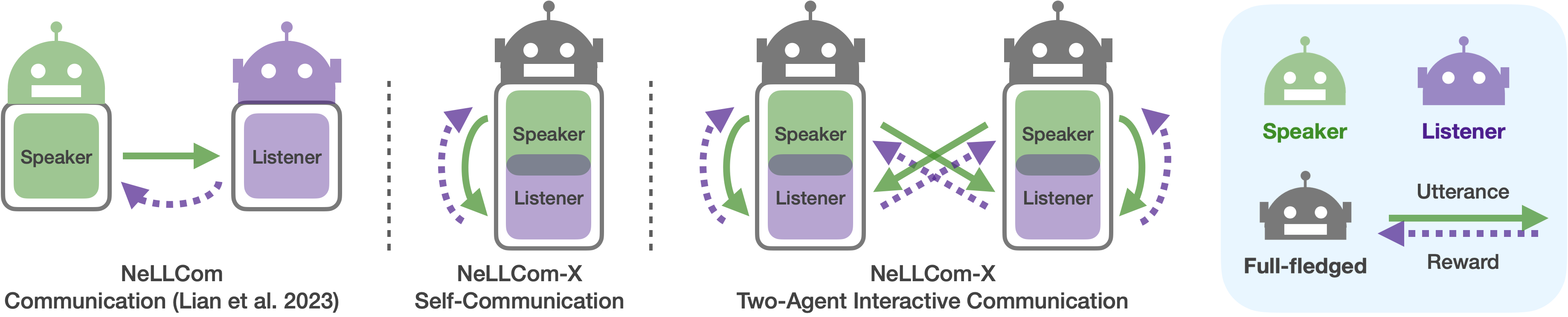}
    \caption{Overview of the NeLLCom-X framework.}
    \label{fig:comm_diagram}
\end{figure*}

However, neural agent emergent communication frameworks could also be a valuable tool to simulate the evolution of more specific aspects of language. Studies with human participants 
have addressed many other aspects such as specific syntactic patterns like word order or morphology \cite{saldana2021regularization, culbertson2012learning, christensen2016environmental, motamedi2022improvisation}, a tendency to reduce dependency lengths \cite{fedzechkina2018human, saldana2021cross}, colexification patterns and the role of iconicity or metaphor in the emergence of new meanings \cite{karjus2021conceptual, verhoef2015emergence, verhoef2016iconicity, verhoef2022interaction, tamariz2018interactive}, and combinatorial organisation of basic building blocks \cite{roberts2012emergence, verhoef2012origins, verhoef2014emergence}.
What most of these studies have in common is that participants are asked to learn and/or interact with \textit{pre-defined artificial languages} specifically designed by the experimenters to study the linguistic property of interest. 
However, the existing neural-agent communication frameworks (often based on EGG \cite{kharitonov-etal-2019-egg}), 
do not enable training agents on pre-defined languages.
A different body of work has studied the \textit{learnability} by neural networks of various types of artificial languages  
\citep{lupyan2002case,wang-eisner-2016-galactic,bisazza-etal-2021-difficulty,white-cotterell-2021-examining,hopkins-2022-towards,kallini2024mission}.
This paradigm has led to important insights, revealing inductive biases of neural models, but is limited to studying learnability in a passive supervised learning setting, unlike the dynamic and interactive setting in which human language has evolved.

A framework combining agent communication with the ability to learn pre-defined artificial languages was recently introduced by \citet{lian-etal-2023-communication}. In NeLLCom (Neural agent Language Learning and Communication), agents are first trained on an initial language through Supervised Learning, followed by a communication phase in which a speaking and listening agent continue learning together through Reinforcement Learning by optimizing a shared  communicative reward.

In this paper, we extend NeLLCom with group interaction with the aim of studying the interplay between learnability of specific pre-defined languages, communication pressures, and group size effects under the same framework.
To this end, we first extend the vanilla NeLLCom agent to act as both listener and speaker (i.e. role alteration, cf. \Cref{fig:comm_diagram}), which was identified as an important gap in the emergent communication literature by \citet{galke2022emergent}.
Then, we design a procedure to let such `full-fledged' agents interact in pairs with either similar or different initial language exposure, or in groups of various sizes. With the extended framework, NeLLCom-X, we replicate the key findings of \citet{lian-etal-2023-communication} and additionally show that (i) pairs of agents trained on different initial languages quickly adapt their utterances towards a mutually understandable language, (ii) languages used by agents in larger groups become more optimized and less redundant, and (iii) a word-order/case-marking trade-off emerges not only in individual speakers, but also at the group level.


%
%
We release NeLLCom-X to promote simulations of other language aspects where interaction and group dynamics are expected to play a key role.\footnote{https://github.com/Yuchen-Lian/NeLLCom-X}





\section{Related Work}
\label{sec:relwork}

\paragraph{Role-alternating agents} 
Initially, most work on emergent communication modeled agents to fulfill separate, complimentary roles (i.e. one agent always speaks, the other always listens). Human language users are, of course, able to take both roles. When listing a set of "design features" of human language, \citet{hockett1960origin} refered to \textit{interchangeability} as the ability of language speakers to reproduce any linguistic message they can understand. In experiments with humans communicating via artificial languages, participants also usually take turns being the speaker and listener \cite{kirby2015compression, namboodiripad2016measuring, roberts2012emergence, verhoef2015emergence, verhoef2022interaction}. Therefore, \citet{galke2022emergent} named role-alternation as a missing key ingredient to close the gap between outcomes of simulations and findings from human language evolution data.  



Exceptions to this trend include the role-alternating architectures of
\citet{kottur2017natural},
\citet{harding-graesser-etal-2019-emergent}, and
\citet{taillandier2023neural}.
Recently, \citet{michel2023revisiting} 
propose a method to couple a speaker and listener among a group of speaking and listening agents. By what they call "partitioning", the listener-part is only trained to adapt to its associated speaker, 
while the listener parameters are frozen during communication with other speakers. Hence, the speaking and listening parts of an agent are tied softly, i.e. no "physical" link via shared modules. 
While being workable, this partitioning seems less realistic in terms of cognitive plausibility and communication, as human listeners continually refine their understanding during all kinds of interactions (speaking as well as listening). 
%
\EXTENSION{ MORE RELATED WORK
\citet{taillandier2023} investigated whether turn-taking could emerge spontaneously in emergent communication when agents could decide when to speak or listen, and speaking at the same time was penalized. Agent pairs that developed better turn-taking achieved higher accuracy at a cooperative game.}
What all these studies have in common is their focus on protocols emerging from scratch, i.e. starting from random symbols, which does not allow for simulations with pre-defined languages.
Closer to our goal, \citet{chaabouni-etal-2019-word} train agents on artificial languages and observe them drift in a simple iterated learning setup that does not model communication success. They use sequence-to-sequence networks that can function both as speaker and listener by representing both utterances and meanings as sequences and merging meaning and word embeddings into a single weight matrix, tied between input and output. 

We combine elements of the above techniques to design agents that can learn artificial languages and use them to interact in a realistic manner.




\paragraph{Group communication}
Natural languages typically have more than two speakers, and language structure is shaped by properties of the population. According to the Linguistic Niche hypothesis, for example, languages used by larger communities tend to be simpler than those used in smaller, more isolated groups \cite{wray2007consequences, lupyan2010language}. Similarly, experiments with human participants have shown that interactions in larger groups can result in more systematic languages \cite{raviv2019larger}. Various emergent communication simulations have been designed to investigate group effects, revealing the emergence of natural language phenomena. 
\citet{tieleman2019shaping}, for example, found that representations emerging in groups are less  idiosyncratic and more symbolic. They model a population of community-autoencoders and since the identities of the encoder and decoder are not revealed within a pair, the emerging representations develop in such way that all decoders can use them to successfully reconstruct the input, resulting in a more simple language as also found in humans. 
\citet{michel2023revisiting} found that larger agent groups develop more compositional languages. 
\citet{harding-graesser-etal-2019-emergent} investigated various language contact scenarios with populations of agents that have first developed distinct languages within their own groups, and could observe the emergence of simpler 'creole' languages, resembling findings from human language contact. 
\citet{kim2021emergent} 
vary the connectivities between agents in groups, and find the spontaneous emergence of linguistic dialects in large groups with over a hundred agents having only local interactions.
Again, none of these frameworks support training agents on pre-defined languages, limiting the extent to which they can be applied to specific human-like linguistic features. 

In this work, we showcase how NeLLCom-X agents can interact in groups using artificial languages that were specifically designed to study the emergence of word-order/case-marking patterns.


\section{NeLLCom-X}
We summarize the original NeLLCom framework \cite{lian-etal-2023-communication} and then explain how we extend it with role alternation and group communication.

\subsection{Original Framework}
NeLLCom agents exchange simple meanings using pre-defined artificial languages.
To achieve this, the framework combines: 
(i) a supervised learning (SL) phase, during which agents are taught a language with specific properties, and 
(ii) a reinforcement learning (RL) phase, during which agent pairs interact 
via a meaning reconstruction game.

\textbf{Meanings} are triplets $m=\{A,a,p\}$ representing simple scenes with an action, agent, and patient, respectively (e.g. \textsc{praise, fox, crow}).
An artificial \textbf{language} defines a mapping between any given meaning $m$ and utterance $u$ which is a variable-length sequence of symbols from a fixed-size vocabulary (e.g. \textit{`Fox praises crow'}). According to the language design, the same meaning may be expressed by different utterances, and vice versa, the same utterance may signal different meanings.

The \textbf{speaking} function \mbox{$\mathcal{S}: m \mapsto u$} is implemented by a linear-to-RNN network, whereas the \textbf{listening} function \mbox{$\mathcal{L}: u \mapsto m$} is implemented by a symmetric RNN-to-linear network.%
\footnote{To make the two networks fully symmetric, we slightly modify the original listener architecture of \citet{lian-etal-2023-communication} by adding a meaning embedding layer before the final softmax. Preliminary experiments show no visible effect on the results.}
The sequential components are implemented as a single-layer Gated Recurrent Unit \cite{chung2014empirical}.
In both directions, meanings are represented by unordered tuples instead of sequences to avoid any ordering bias, differently from \citet{chaabouni-etal-2019-word} who also represent meanings as sequences.

The \textbf{SL} phase minimizes the cross-entropy loss of the predicted words given meaning (speaker) or the predicted meaning tuple given utterance (listener) with respect to a gold-standard dataset $D={(m,u)}$.
The \textbf{RL} phase maximizes a shared reward $r(m,\hat{u})$ evaluated by the listener’s prediction  $\mathcal{L}(\hat{u})$ given the speaker-generated utterance $\hat{u}=\mathcal{S}(m)$. 
More details on the SL and RL procedures, the respective training objectives, and network architectures are given in~\Cref{app:nellcom-details}.

Crucially, each agent in the original NeLLCom can either function as listener (utterance-to-meaning) or as speaker (meaning-to-utterance), but not as both, see \Cref{fig:comm_diagram}. 
While this minimal setup was sufficient to simulate the emergence of the word-order/case-marking trade-off \cite{lian-etal-2023-communication}, it does not allow for role alternation --a missing key ingredient for realistic simulations of emergent communication \cite{galke2022emergent} and a necessary condition to simulate group communication.



\subsection{Full-fledged Agent}
\label{sec:method_fullAgent}

To realize a full-fledged agent ($\alpha$) that can speak \textit{and} listen while interacting with other agents, we pair two networks \mbox{$\alpha_{i}=(N_{i}^\mathcal{S},N_{i}^\mathcal{L})$}
using two strategies: parameter sharing and self-play (Fig. \ref{fig:comm_diagram}).


\paragraph{Parameter sharing}
A common practice in NLP is tying the weights of the embedding (input) and softmax (output) layers to maximize performance and reduce the number of parameters in large language models \cite{press-wolf-2017-using}. 
\citet{chaabouni-etal-2019-word} applied this technique to their sequence-to-sequence utterance$\leftrightarrow$meaning architecture.
However in our setup, listening and speaking are implemented by two separate, symmetric networks. We then tie the input embedding of the speaking network to the output embedding of the listening network $\mathbf{X}(N_i^\mathcal{S}) = \mathbf{O}(N_i^\mathcal{L})$ (both representing meanings). 
Likewise, we tie the input embedding of the listener to the output embedding of the speaker $\mathbf{X}(N_i^\mathcal{L}) = \mathbf{O}(N_i^\mathcal{S})$ (both representing words). 
Because of these shared parameters, the speaker training process will also affect the listener, and vice versa. 
To balance listener and speaker optimization during supervised learning, we alternate between the two after each epoch.%
\footnote{As verified in preliminary experiments, results are similar whether the last epoch is a listening or speaking one. } 

\paragraph{Self-play}
Even when word and meaning representations are shared, the rest of the speaking and listening networks remain disjoint, potentially causing the speaking and listening abilities to drift in different directions.
As discussed in \Cref{sec:relwork}, a realistic full-fledged agent should be able to understand itself at any moment.
To ensure this, we let the agent's speaking network send messages to its own listening network while optimizing the shared communicative reward $r$, a procedure known as self-play in emergent communication literature \citep{lowe2019interaction,lazaridou-etal-2020-multi}.
In \Cref{sec:exp_diffLangs}, we show empirically that self-play is indeed necessary to preserve the agents' self-understanding while their language evolves in interaction.



\subsection{Interactive Communication}
\label{sec:method_interact}
Given the new full-fledged agent definition, communication becomes possible between two or more role-alternating agents.
We introduce the notion of \textit{turn} to denote a minimal communication session where RL weight updates take place between an agent's speaker and either its own listener or another agent's listener:
\begin{align}
\mathsf{self\_turn}(\alpha_i) &=\mathsf{RL}(N^\mathcal{S}_i,N^\mathcal{L}_i) \\
\mathsf{inter\_turn}(\alpha_i,\alpha_j) &=\mathsf{RL}(N^\mathcal{S}_i,N^\mathcal{L}_j) 
\end{align}
For example, in our experiments, a turn corresponds to 10 batches of 32 meanings. 
%
%
\EXTENSION{
\begin{multline}
\mathsf{self-Comm}(\alpha_i) = \mathsf{Iter} (\mathsf{self\_update}(\alpha_i))
\end{multline}
\begin{multline}
 \mathsf{Interactive-Comm}(\alpha_i,\alpha_j) = \mathsf{Iter}(\\
 \mathsf{inter\_update}(\alpha_i,\alpha_j), \mathsf{inter\_update}(\alpha_j,\alpha_i),\\
\mathsf{self\_update}(\alpha_i), \mathsf{self\_update}(\alpha_j))
\label{fuc:lst_loss}
\end{multline} 
\begin{multline}
 \mathsf{Group-Comm}(Agents) = \mathsf{Iter}(\\
 \mathsf{inter\_update}(\alpha_i,\alpha_j),\; \forall (\alpha_i \to \alpha_j) \in Comm\_graph\\
\mathsf{self\_update}(\alpha_k), \; \forall \alpha_k \in Agents)
\label{fuc:lst_loss}
\end{multline} 
}
Note that interaction can involve agents that were trained on the same language, or on different initial languages, as we will show in \Cref{sec:exp_interact}.


\begin{algorithm}[b!]
  \small
  \caption{Group Communication}
  \label{algo:groupComm}
  \LinesNumbered
  \KwIn{set of SL-trained agents: $Agents$,\\
  \Indp edges in the connectivity graph: $\mathcal{G}$,\\
  $n\_rounds$, $\sigma$}
  \For{$r=1:n\_rounds$}{
  $comm\_turns =$ shuffle$(\mathcal{G})$\\
  \For{$turn_i \in comm\_turns$}
  {
      $i_{spk},\; i_{lst}$ = $turn_i$\\
      $\alpha_{spk}$ = $Agents$[$i_{spk}$],$\;$
      $\alpha_{lst}$ = $Agents$[$i_{lst}$]\\
      inter\_turn($\alpha_{spk}$, $\alpha_{lst}$)\\
      \For{$\alpha=\{\alpha_{spk}, \alpha_{lst} \}$}
      {
      $\alpha$.activation += 1\\
      \If{$\alpha$.activation >= $\sigma$}
          {self\_turn($\alpha$)\\
          $\alpha$.activation = 0}
      }
  }}
  \end{algorithm}
  
\paragraph{Turn scheduling}
During group communication, a connectivity graph $\mathcal{G}$ is used to define which agents can communicate with another, and which cannot. Within $\mathcal{G}$, a node $i$ represents an agent and a directed edge ($i,j$) represents a connection whereby $\alpha_i$ can speak to $\alpha_j$, but not necessarily vice versa.
%
%
Turn scheduling then proceeds as shown in Algorithm~\ref{algo:groupComm}:
Before each turn, an edge $(i,j)$ is sampled without replacement from $\mathcal{G}$.
Then $\alpha_i$ and $\alpha_j$ perform an $\mathsf{inter\_turn}$ of meaning reconstruction game, 
with $\alpha_i$ acting as the speaker and $\alpha_j$ as the listener.
Interactive turns are interleaved with self-play turns at fixed intervals, i.e. every time an agent has participated in $\sigma \times \mathsf{inter\_turn}$, it performs one $\mathsf{self\_turn}$.
Once all edges in $\mathcal{G}$ have been sampled, a communication \textit{round} is complete.
In this work, we only consider a setup where all agents can interact with all other agents ($\mathcal{G}$ is a complete directed graph). 
We leave an exploration of more complex configurations such as those studied by \citet{harding-graesser-etal-2019-emergent,kim2021emergent,michel2023revisiting} to future work.
We set $\sigma=10$ in all interactive experiments, unless differently specified. 
Interaction between two agents follows the same procedure as group communication.

\section{Experimental Setup}

As our use case, we adopt the same artificial languages as \citet{lian-etal-2023-communication}. These simple verb-final languages vary in their use of word order and/or case marking to denote subject and object, and were originally proposed by \citet{fedzechkina2017balancing} to study the existence of an effort-informativeness trade-off in human learners.

\paragraph{Artificial languages}
The meaning space includes 10 entities and 8 actions, resulting in a total of 10$\times$(10$-$1)$\times$8$=$720 possible meanings. 
\EXTENSION{(an entity cannot be agent and patient at the same time)}
Utterances can be either SOV or OSV. The order profile of a language is defined by the proportion of SOV, e.g. 100\% fixed, 80\% dominant, 50\% maximally flexible-order. 
Objects are optionally followed by a special token `mk' while subjects are never marked. 
To simplify the vocabulary learning problem, each meaning item correspond to exactly one word, leading to a vocabulary size of 10$+$8$+$1$=$19.
Two example languages are shown in Table~\ref{tab:language-table}.

\begin{table}[h!t]
\begin{center} \small
\begin{tabular}{@{\ \ }c@{\ \ } | @{\ \ }c@{\ \ } | @{\ \ }c@{\ \ }}
\hline 
\bf language & \bf properties & \bf possible utterances \\
\hline
100s+0m & 100\% \textsc{sov}; 0\% marker & \it Tom Jerry chase \\
\hline
\multirow{2}{*}{80s+100m} & 80/20\% \textsc{sov/osv} &\it Tom Jerry mk chase \\
& 100\% marker &\it Jerry mk Tom chase \\
\hline
\end{tabular}
\end{center}
\caption{\label{tab:language-table} Two example languages with varying order and marking proportions, and corresponding utterances for the meaning $m$=\{$A$: \textsc{chase}, $a$: \textsc{tom}, $p$: \textsc{jerry}\}. 
}
\end{table}

\EXTENSION{Illustrate optional marker by making first example: 100s+50m instead of 100s+0m}



\paragraph{Evaluation}
Following \citet{lian-etal-2023-communication}, agents are evaluated on a held-out set of meanings unseen during any training phase.
\EXTENSION{
During SL, we measure listening and speaking accuracy: these metrics consider a meaning or utterance correct only if they fully match those in the gold-standard dataset $D$.
By contrast, \textit{permissive} speaking accuracy is 1 if a generated utterance fully matches \textit{any} of the possible utterances admitted for a given meaning by the grammar. 
}
The SL phase is evaluated by listening/speaking accuracy computed against gold dataset $D$, while the RL phase is evaluated by meaning reconstruction accuracy, or communication success. 
In NeLLCom-X, communication success denotes two different aspects: self-understanding when measured between the same agent's speaker and listener network, or interactive communication success when measured between a speaking agent and a different listener agent:
\begin{align}
acc_{self}(m, \alpha_i) &= acc(m, \mathcal{L}_{\alpha_i}(\mathcal{S}_{\alpha_i}(m)) \\
acc_{inter}(m, \alpha_i, \alpha_j) &= acc(m, \mathcal{L}_{\alpha_j}(\mathcal{S}_{\alpha_i}(m))
\end{align}
\EXTENSION{
\[
acc(m, \hat{m}) = \begin{cases}
1, & \text{if } m_{e} = \hat{m_{e}},\; \forall e \in \{A, a, p\} \\
0, & \text{otherwise}
\end{cases}
\]
}
where $acc(m, \hat{m})$ is 1 iff the entire meaning is matched.
Interactive success is not symmetric.
 




\paragraph{Production preferences}
Besides accuracy, our main goal is to observe \textit{how} the properties of a given language evolve throughout communication. 
This is done by recording the proportion of markers and different orders in a set of utterances generated by an agent for a held-out meaning set, after filtering out utterances that are not recognized by the initial grammar.
When the focus is on the trade-off, rather than on a specific word order, we measure \textit{order entropy}.
Production preferences can be aggregated over an individual agent, a group, or the entire population.

\EXTENSION{
\lyc{Inspiring by the Uncertainty versus production effort trade-off plot from \citet{fedzechkina2017balancing}, we come up with a more straightforward way to visualize the trade-off between word-order and case marking, i.e., Word-order Entropy versus Marker proportion plot \ref{fig:color}.
\footnote{One shortage of Uncertainty-Effort plot is that in case of high effort (i.e., majority of marker) with close to zero uncertainty, it's not clear whether order flexibility is maintained.}
Different corners represent different word-order - case-marking trade-off conditions. Specifically, [green] at the top-left represents fixing order fully mafrked language, the redundant condition, [yellow] at the top-right represents language with fully marked, flexible (50-50) order language, [blue] at the bottom-left represents fixing-order no marker language, and [red] at the bottom right represents flexible-order without marking language which is also the ambiguous condition. We also apply this visualization to all miniature languages used in later experiments, shown as different diamonds in the plot.}
\begin{figure}[h]
\centering
\captionsetup{singlelinecheck=off}
  \includegraphics[width=\columnwidth]{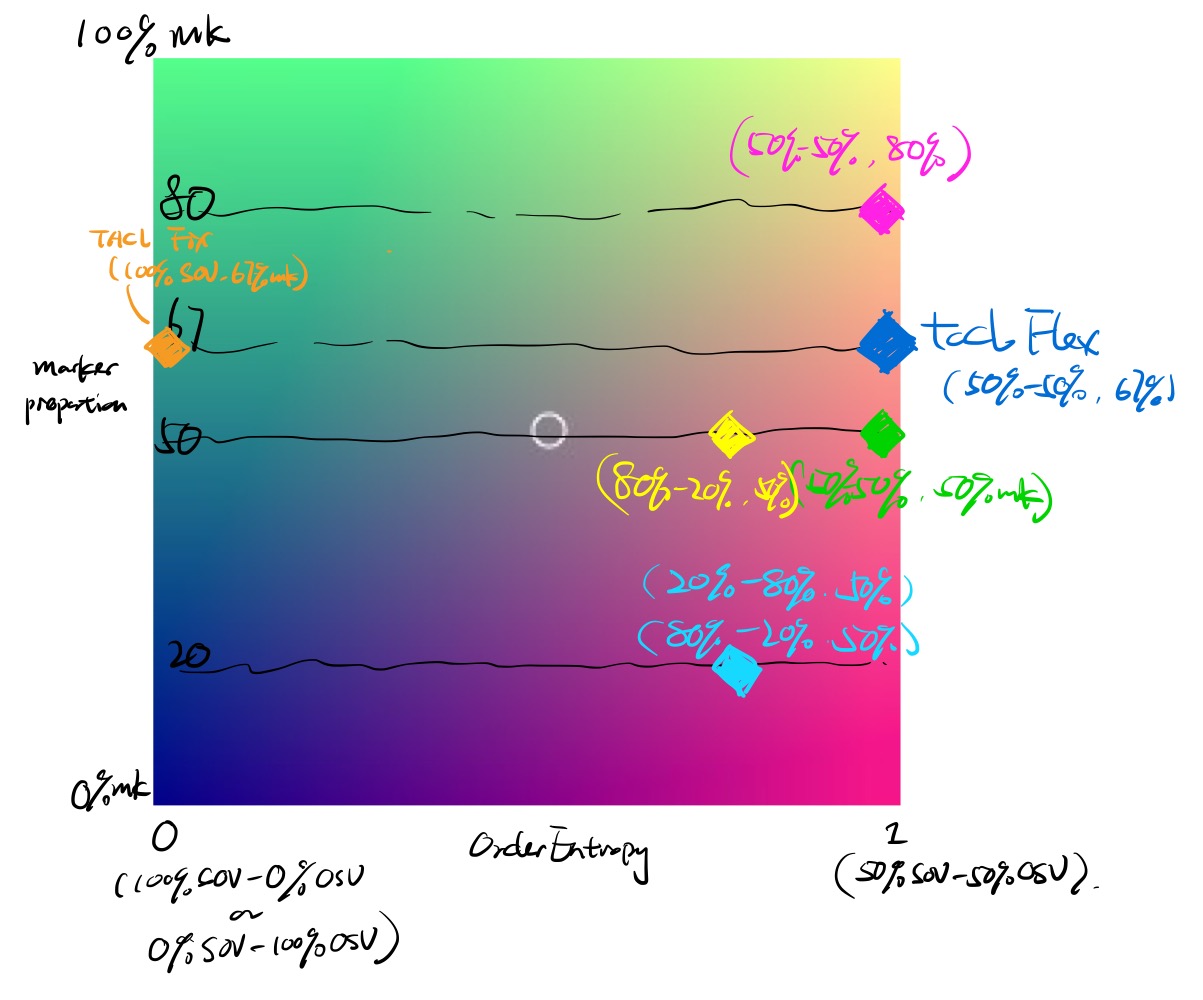}
\caption{
marker/orderentropy plot \& all predefined languages
}
\label{fig:color}
\end{figure}
}

\EXTENSION{
\paragraph{Jensen-Shannon Divergence}
}


\section{Replicating the Trade-off with Full-fledged Agents}
\label{sec:exp_selfComm}

Before moving to interactive communication,
we validate the new NeLLCom-X framework through a replication of \citet{lian-etal-2023-communication}'s main findings.
The simple speaker-listener communication setup of NeLLCom could be seen as a speaker-internal monitoring mechanism predicting the utterance understandability \cite{ferreira2019mechanistic}. 
\EXTENSION{
\TV{I would suggest to remove this interpretation of Fedz results since a paper by Smith and Culbertson suggest a different explanation (based on iconicity, but no space to discuss it here) and what is happening in the experiment is slightly different from what the Ferreira paper talks about, which is audience design and adaptation to a specific interlocuter... so even though I do think that this plays a role in Fedz, I think it needs more elaborate explanation/discussion}
}
Here, we compare NeLLCom results to those of NeLLCom-X full-fledged agents only engaging in self-play.
We use SL to train two sets of agents on the exact same languages as \citet{lian-etal-2023-communication}, respectively: 100s+67m for fixed-order and 50s+67m for flexible-order. Then, every agent performs 60 $\mathsf{self\_turn}$ iterations causing its production preferences to drift.

After SL, our agents have successfully learnt both languages but no regularization happens, as expected. 
By contrast, the results of self-play averaged over each 50-agent set indicate that both languages progressively lose markers. 
Crucially, the fixed-order language does so faster than the flexible one, where markers are often necessary for agent/patient disambiguation.
In sum, self-play in NeLLCom-X results in very similar trends as the simple NeLLCom setup, confirming the emergence of a human-like order/marking trade-off \cite{fedzechkina2017balancing}.
Detailed replication results are provided in \Cref{app:replicateTacl}. 
Here, we report communication success during self-play and production preferences at the end of self-play for the flexible language (\Cref{fig:replicate}, top row).
Self-understanding increases through RL leading to a much more informative language, while production preferences reveal that this spans from an overall decrease in order entropy with marking proportion remaining almost the same on average (solid circle). 
While some agents approach the optimal points of fixed-order/no-marking (bottom-left corner) or flexible-order/full-marking (top-right), the large variability in production preferences suggests many agents settle on less optimized, redundant languages, as also found by \citet{lian-etal-2023-communication}.


\begingroup
\setlength{\tabcolsep}{0pt} 
\renewcommand{\arraystretch}{0.6}
\begin{table}[t]
  \centering
  \begin{tabular}{c c c}

  & \shortstack{\small Communicative success \\ \small per turn}
  &
  \shortstack{\small Marker use \\ \small by order entropy}\\

  \rotatebox[origin=c]{90}{\small 50s+67m} &
  \begin{minipage}{.24\textwidth}
      \includegraphics[width=\columnwidth]{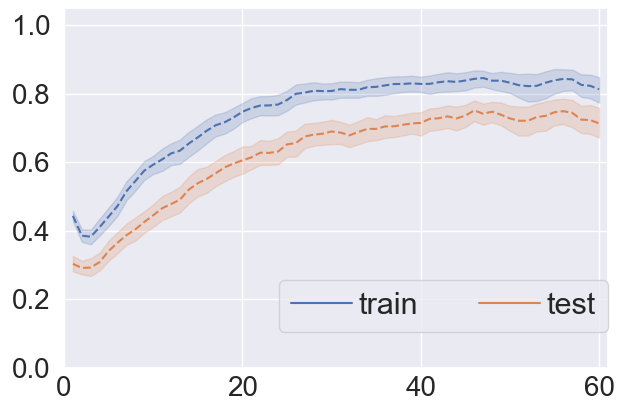}
    \end{minipage}
    &
    \begin{minipage}{.24\textwidth}
      \includegraphics[width=\columnwidth]{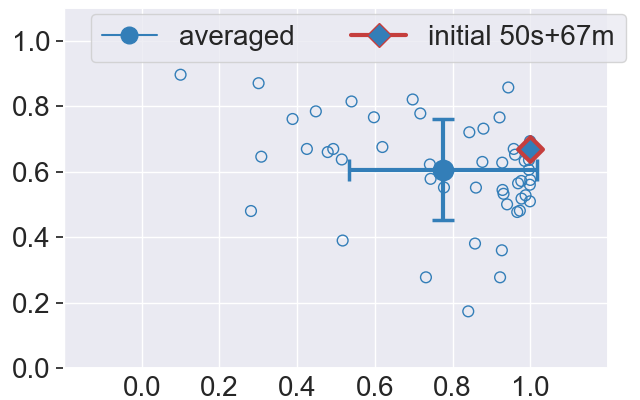}
    \end{minipage}
    \\

    \rotatebox[origin=c]{90}{\small 50s+50m} &
  \begin{minipage}{.24\textwidth}
      \includegraphics[width=\columnwidth]{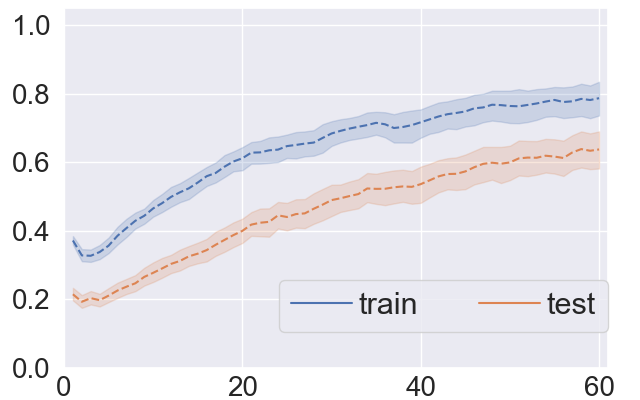}
    \end{minipage}
    &
    \begin{minipage}{.24\textwidth}
      \includegraphics[width=\columnwidth]{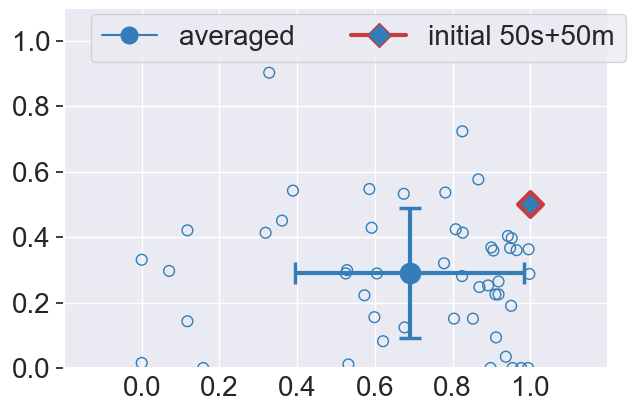}
    \end{minipage}

  \end{tabular}
  \captionof{figure}
  {
Two populations of 50 agents engaging in self-play (no interaction) after having learned two flexible-order, optional-marker languages: one with 67\% the other with 50\% marking.
Left column: Average communication success across self-play turns.
Right column: 
Production preferences: solid diamonds mark the initial language; each empty circle denotes a full-fledged agent at the end of self-play; solid circles are the average of all agents, with error bars showing standard deviation.
}
\caption*{}
\vspace{-2em}
\label{fig:replicate}
\end{table}
\endgroup


\paragraph{Initial marking proportion} \label{par:initial_marking}
We reconsider here a language design choice of \citet{lian-etal-2023-communication} who, in turn, inherited it from the human study of \citet{fedzechkina2017balancing}. 
It was recently found that human learners exposed to a fixed-order language with 75\% marking tend to regularize by increasing marker use even though this would make the language less efficient \cite{tal2022impact}. 
Similarly, the dominant proportion (67\%) of marking utterances in our initial languages may push the agents to prefer marking even when it may be a redundant strategy. 
Hence, we propose that a more balanced distribution of 50\% markers and 50/50\% word order may be a better choice to reveal the intrinsic preferences of the learners, if there are any, without biasing them to regularize markers.
Results in \Cref{fig:replicate} (bottom row) show that this language has overall lower communicative success, as expected given the higher amount of ambiguous sentences. 
However, success increases substantially during interaction while production preferences reveal a larger variability in solutions including those with more fixed order and less markers.
We use this more neutral combination as the default language in all remaining experiments.

\section{Interactive Communication Results}
\label{sec:exp_interact}

This section presents our main results:
in \Cref{sec:exp_diffLangs} we focus on pairwise interaction and show how NeLLCom-X can be used to simulate communication between speakers of different languages, which was not possible in the original framework;
in \Cref{sec:exp_groupSize} we move to group communication and study the effect of group dynamics on communication success and production preferences.
Training details for this section are given in \Cref{app:inter_traindetail}.

\subsection{Speakers of Different Languages}\label{sec:exp_diffLangs}


We study a simple setup with two full-fledged agents interacting with each other in both ways $\alpha_{base}$$\leftrightarrow$$\alpha_{other}$. 
The first ($\alpha_{b}$ for \textit{base}) is always trained on the neutral language 50s+50m, while the second ($\alpha_{o}$ for \textit{other}) is trained on one of four languages with different properties. 
If interaction works, we expect (i) agent pairs to negotiate a mutually understandable language and (ii) $\alpha_{b}$'s language to drift in different directions according to its interlocutor.
\EXTENSION{We repeat each experiment with 50 random seeds (i.e. 50 agent pairs).}
For production preferences, we are interested here in the specific word order of the evolving languages so we plot proportion of markers against \textit{proportion of SOV} instead of order entropy. 


\begingroup
\setlength{\tabcolsep}{0pt} 
\renewcommand{\arraystretch}{0.8}
\begin{table}[h!]
  \centering
  \begin{tabular}{l c c}

& \shortstack{\small Communicative success \\ \small per turn}
  &
  \shortstack{\small Marker use \\ \small by order (SOV/all)}\\

  \rotatebox[origin=c]{90}{\small $\alpha_{b}\leftrightarrow\alpha_{b}$} &
  \begin{minipage}{.24\textwidth}
      \includegraphics[width=\columnwidth]{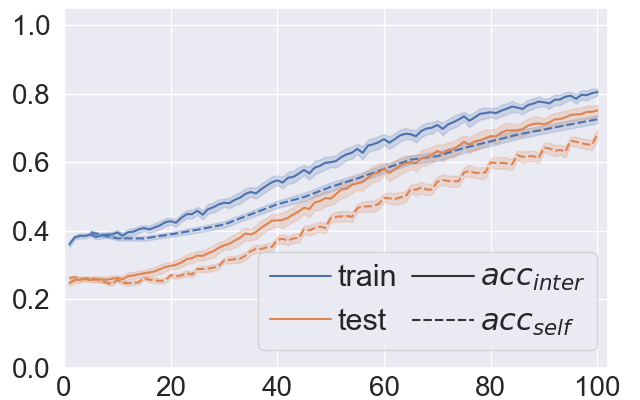}
    \end{minipage}
    &
    \begin{minipage}{.24\textwidth}
      \includegraphics[width=\columnwidth]{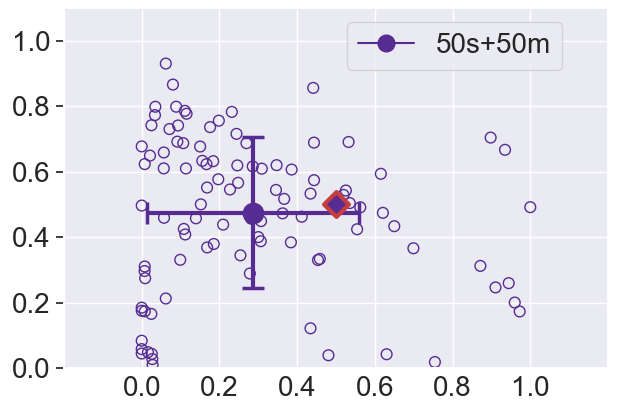}
    \end{minipage}
    \\

    \rotatebox[origin=c]{90}{\small $\alpha_{b}\leftrightarrow$ 80s+20m} 
    &
    \begin{minipage}{.24\textwidth}
    \includegraphics[width=\columnwidth]{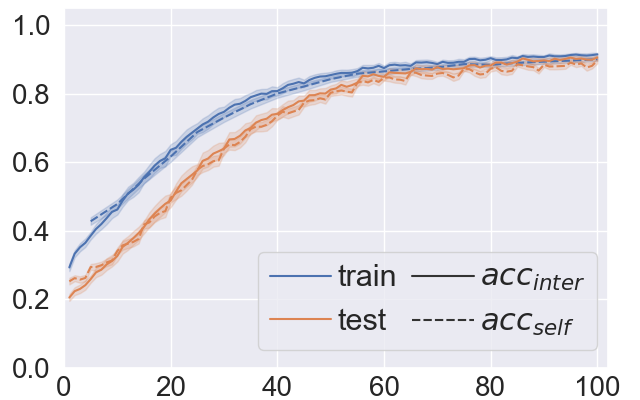}
    \end{minipage}
    &
    \begin{minipage}{.24\textwidth}
      \includegraphics[width=\columnwidth]{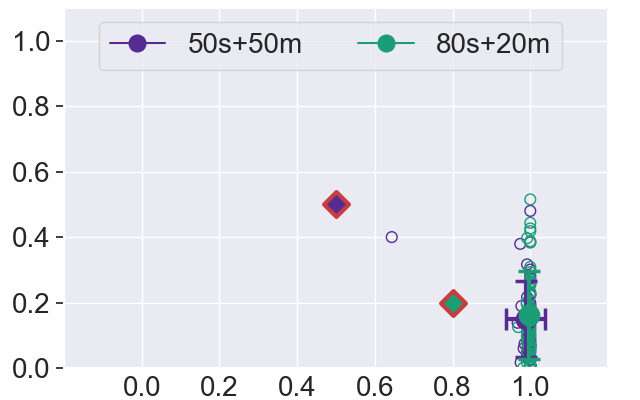}
    \end{minipage}
    \\

    \rotatebox[origin=c]{90}{
    \small $\alpha_{b}\leftrightarrow$ 20s+20m 
    }
    &
    \begin{minipage}{.24\textwidth}
      \includegraphics[width=\columnwidth]{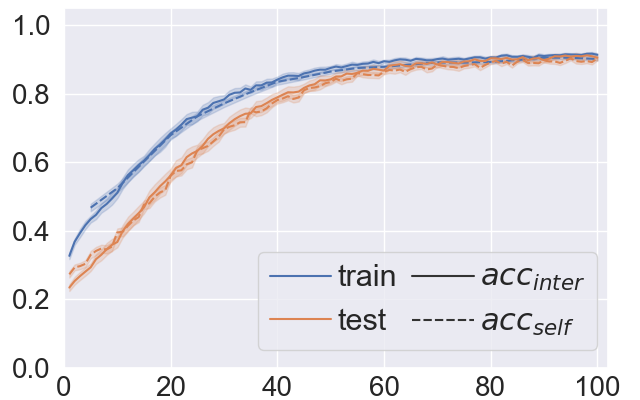}
    \end{minipage}
    &
    \begin{minipage}{.24\textwidth}
      \includegraphics[width=\columnwidth]{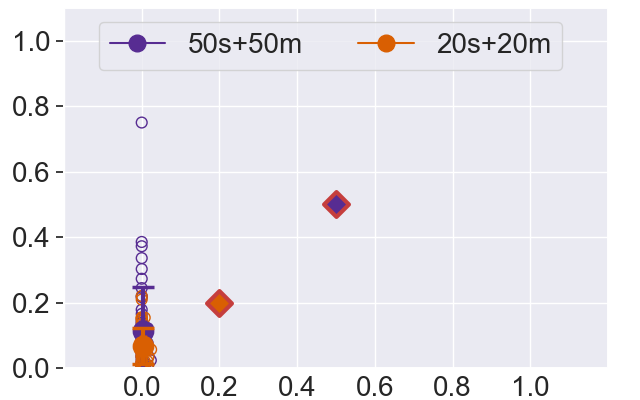}
    \end{minipage}
    \\

    \rotatebox[origin=c]{90}{
    \small $\alpha_{b}\leftrightarrow$ 50s+80m 
    }
    &
    \begin{minipage}{.24\textwidth}
    \includegraphics[width=\columnwidth]{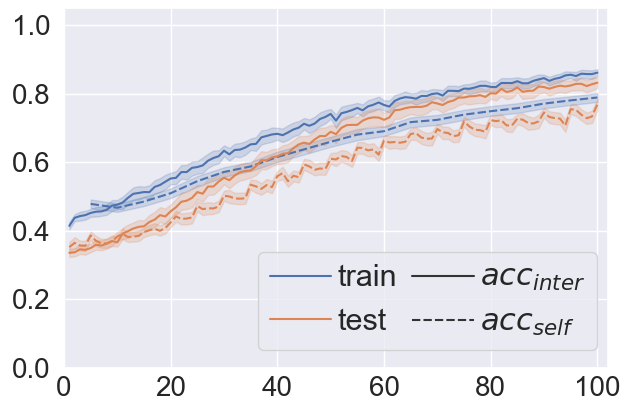}
    \end{minipage}
    &
     \begin{minipage}{.24\textwidth}
      \includegraphics[width=\columnwidth]{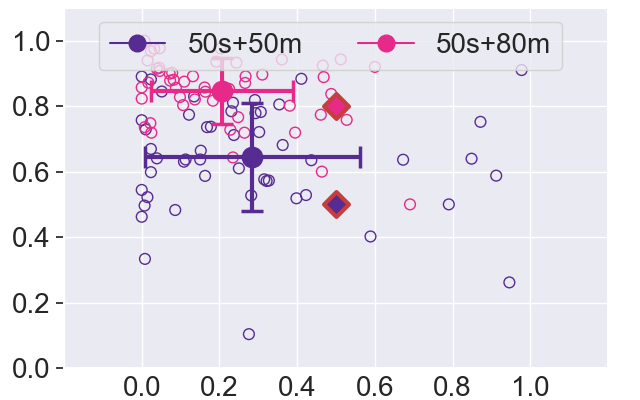}
    \end{minipage}
    \\

    \rotatebox[origin=c]{90}{
    \small $\alpha_{b}\leftrightarrow$ 80s+50m 
    }
    &
    \begin{minipage}{.24\textwidth}
      \includegraphics[width=\columnwidth]{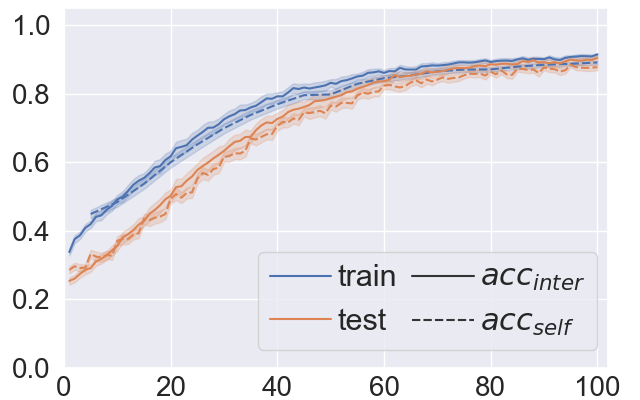}
    \end{minipage}
    &
    \begin{minipage}{.24\textwidth}
      \includegraphics[width=\columnwidth]{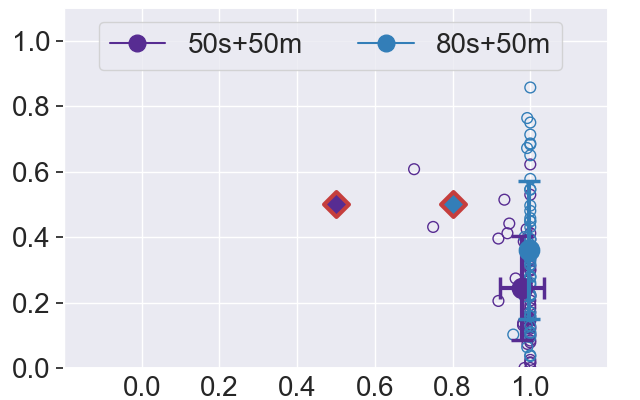}
    \end{minipage}
    \\
    
  \end{tabular}
  \captionof{figure}
  {Interactive communication between different language speakers. The first agent is always trained on 50s+50m ($\alpha_{b}$). Each experiment is repeated with 50 agent pairs.
  }
  \caption*{}
  \vspace{-2em}
  
\label{fig:mix_comm}
\end{table}
\endgroup

The communication success plots in \Cref{fig:mix_comm} (left column) show a faster convergence and higher final accuracy when $\alpha_{o}$ has a stronger order preference.
As for production preferences (\Cref{fig:mix_comm}, right column), 
in the control setting where two neutral agents interact with each other, most agents move towards either side of the plot, representing order regularization. A larger portion of agents regularize towards OSV rather than SOV, which was also observed by \citet{lian-etal-2023-communication} and might be due to OSV being the order where the disambiguating marker appears earlier.
Marking decreases only slightly on average. 
%
The next two settings involve initial languages with few markers and different order preferences but equally low order entropy (20s+20m and 80s+20m). 
As shown by the highly symmetric trends, these pairs strongly converge by regularizing towards the dominant order of $\alpha_{o}$ and further reducing markers.
The fourth setting involves a language where marking is widespread and informative due to high order entropy (50s+80m). 
Here, $\alpha_{b}$ shows on average a similar order regularization as in the control setting $\alpha_{b}$$\leftrightarrow$$\alpha_{b}$, but with a marking increase instead of decrease. 
Finally, when involving a dominant-order language with no clear marking preference (80s+50m), agents strongly regularize the dominant order, with a majority of them reducing marker use.

Taken together, these results demonstrate that (i)~pairs of different-language agents succeed in negotiating a mutually understandable language in most cases, and (ii)~the evolution of an agent's language strongly depends on whom they interact with, thereby matching the expectations for a realistic simulation of interactive communication.

\EXTENSION{
Check (and plot) conditional marking. E.g. by  computing probOSV|marker/probSOV|marker?
}

\paragraph{Impact of self-play during interactions}
\label{sec:self_comm_freq}
As explained in \Cref{sec:method_interact}, each agent performs a turn of self-play after completing $\sigma=10$ turns of interactive communication, based on preliminary experiments.
\EXTENSION{Here we compare this setup to more frequent self-play ($\sigma=2$) and to no self-play at all ($\sigma=\inf$).}
We compare this to a setup where no self-play is performed during interaction ($\sigma=\inf$), in the case where two agents start from a state of poor mutual understanding due to limited marking and strongly diverging order preferences (80s+20m vs. 20s+20m).
%
As shown in \Cref{fig:self-play-effect}, disabling self-play leads to
extremely low self-understanding even though communication \textit{between} the two agents is successful. 
\EXTENSION{
To explain this result, we show in the bottom row the production preferences of a single agent pair (chosen to be representative of the general behavior of the 20 \AB{PLS CONFIRM THIS?}\lyc{here or tab.4 caption?} random seeds). We can see that the two agents regularized their language in opposite directions, indicating a total decoupling of the speaking and listening ability.
}
To explain this result, we inspect the production preferences of individual agent pairs and find that many regularize their language in opposite directions (e.g. dominant SOV vs. dominant OSV, both with no markers), indicating a total decoupling of the speaking and listening ability.
Thus, we confirm that embedding tying alone does not allow for a realistic interaction simulation, making self-play necessary in our framework.

\begingroup
\setlength{\tabcolsep}{0pt} 
\renewcommand{\arraystretch}{0.9}
\begin{table}[bt]
  \centering
  \begin{tabular}{ c c}
  

   \small w/ self-play ($\sigma$$=$10) & \small w/o self-play  ($\sigma$$=$$\inf$)
  \\
    \begin{minipage}{.245\textwidth}
      \includegraphics[width=\columnwidth]{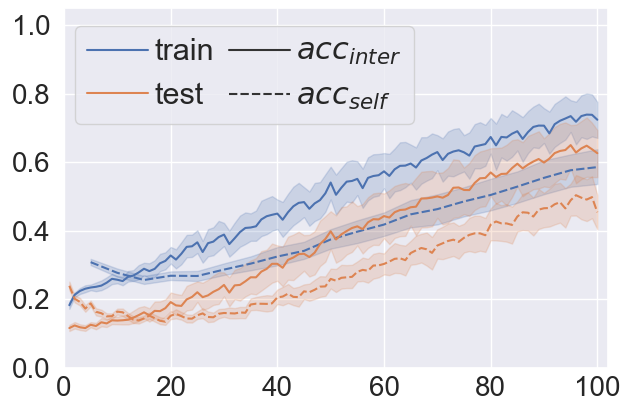}
    \end{minipage}
    &
    \begin{minipage}{.245\textwidth}
      \includegraphics[width=\columnwidth]{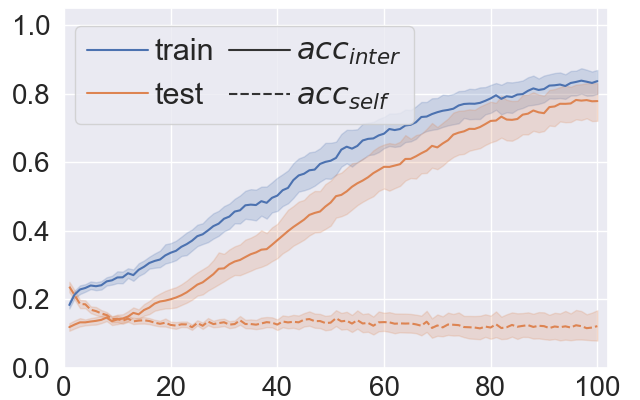}
    \end{minipage}

  \end{tabular}
  \captionof{figure}{
Impact of self-play during interaction in pairs of agents speaking 80s+20m and 20s+20m respectively. Each experiment is repeated with 20 agent pairs, and the average communication per turn is shown. 
}
\caption*{}
  \vspace{-2em}
\label{fig:self-play-effect}
\end{table}
\endgroup

\subsection{Effect of Group Size}\label{sec:exp_groupSize}

Here we move back to a setup where all agents are trained on the same neutral and unstable initial language (50s+50m), but this time they interact in groups of different sizes (2, 4, 8, 20) using the standard self-play frequency ($\sigma=10$).
%
\EXTENSION{
Specifically, we consider four group sizes: 2, 4, 8, and 20 for basic, small, middle, and large groups respectively.
All groups are fully connected.}
To make results comparable, we ensure the \textit{total} number of interactive turns per agent is the same ($\approx$200) in all setups, by setting $comm\_round$ to 100, 34, 15, and 6 respectively. A total of 200 agents are trained in each group size setting.\footnote{100 runs of group of 2, 50 of 4, 25 of 8, and 10 of 20. See all group-specific training details in \Cref{app:inter_traindetail}. In this paper, we only consider fully connected communication graphs and fix the total amount of trained agents to enable comparison. 
We leave an exploration of other group communication factors, such as density and connectivity, to future work.}

\begingroup
\setlength{\tabcolsep}{0pt} 
\renewcommand{\arraystretch}{1.2}
\begin{table}[t!]
  \centering
  \begin{tabular}{l c  c}
  & \shortstack{\small Communicative success \\ \small per turn}
  &
  \shortstack{\small Marker use \\ \small by order entropy}\\

  \rotatebox[origin=c]{90}{\small groups of 2}&
    \begin{minipage}{.24\textwidth}
      \includegraphics[width=\columnwidth]{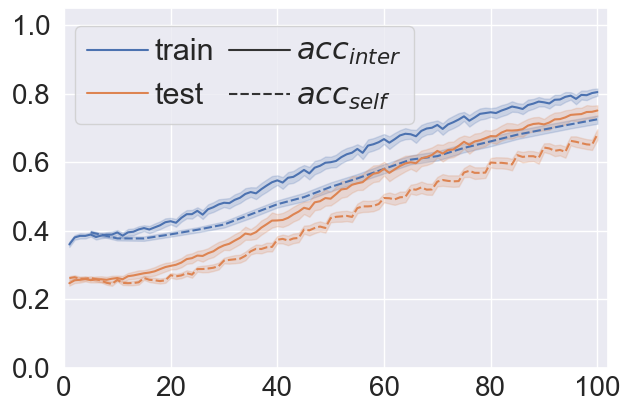}
    \end{minipage}
    &
    \begin{minipage}{.24\textwidth}
      \includegraphics[width=\columnwidth]{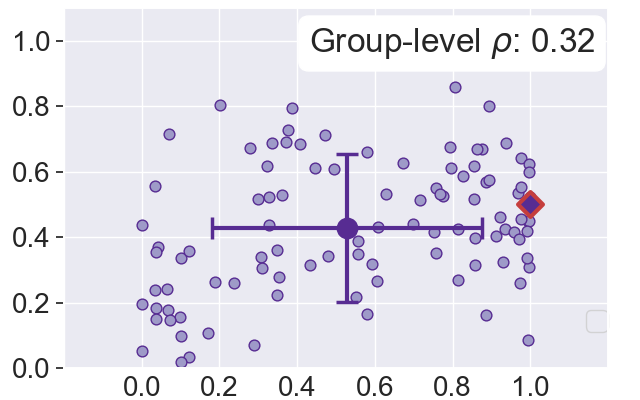}
    \end{minipage}
    \\

    \rotatebox[origin=c]{90}{\small groups of 4}&
    \begin{minipage}{.24\textwidth}
      \includegraphics[width=\columnwidth]{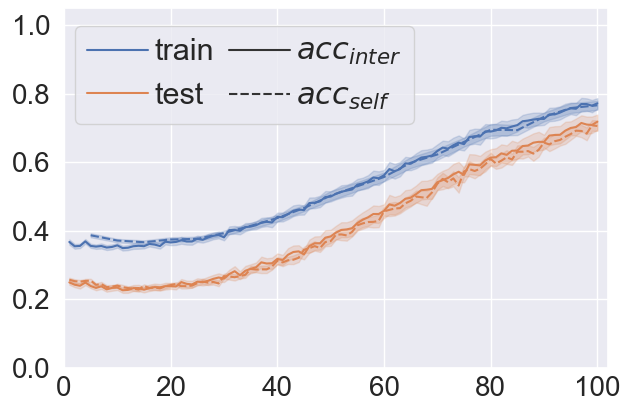}
    \end{minipage}
    &
    \begin{minipage}{.24\textwidth}
      \includegraphics[width=\columnwidth]{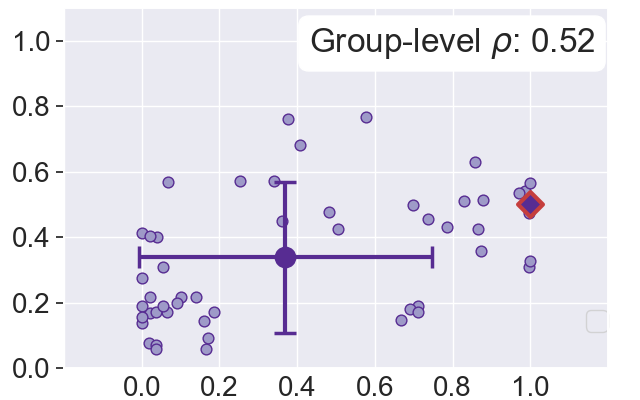}
    \end{minipage}
    \\

    \rotatebox[origin=c]{90}{\small groups of 8}&
    \begin{minipage}{.24\textwidth}
      \includegraphics[width=\columnwidth]{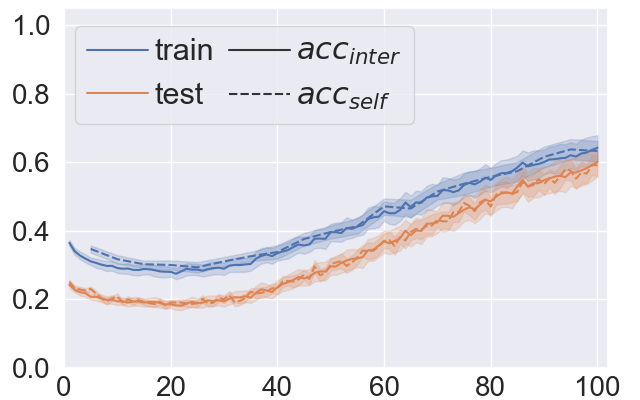}
    \end{minipage}
    &
    \begin{minipage}{.24\textwidth}
      \includegraphics[width=\columnwidth]{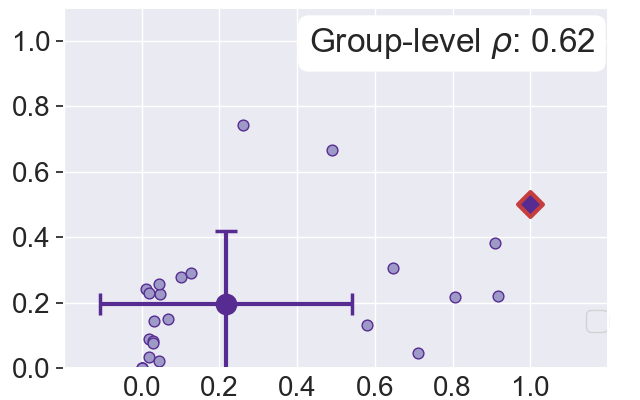}
    \end{minipage}
    \\

    \rotatebox[origin=c]{90}{\small groups of 20}&
    \begin{minipage}{.24\textwidth}
      \includegraphics[width=\columnwidth]{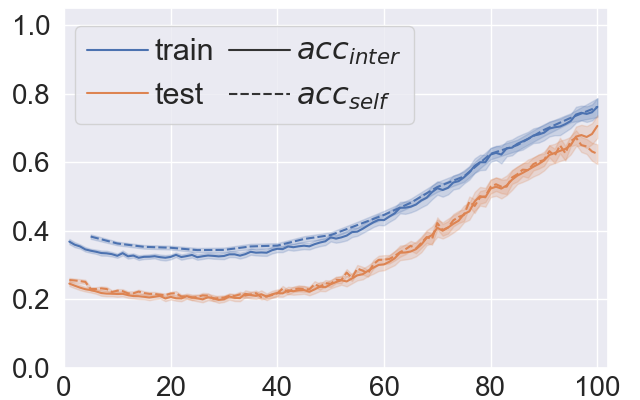}
    \end{minipage}
    &
    \begin{minipage}{.24\textwidth}
      \includegraphics[width=\columnwidth]{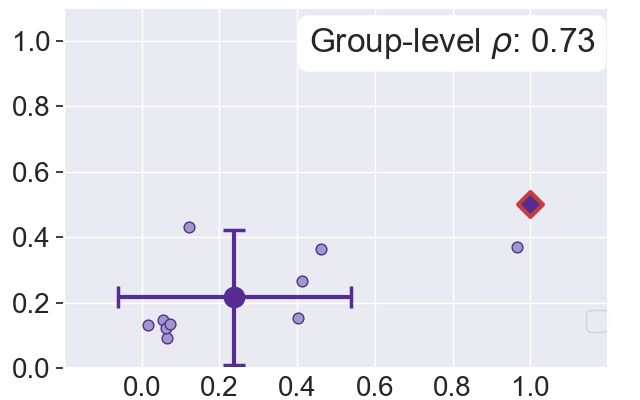}
    \end{minipage}
    
  \end{tabular}
\captionof{figure}
{
Interactive communication in groups of same-language speakers (50s+50m).
Right column: Group-level production preferences (each point is a group) and Spearman's correlation $\rho$ between marker use and order entropy.
}
\caption*{}
\vspace{-2em}
\label{fig:group-results}
\end{table}
\endgroup

\Cref{fig:group-results} (left column) shows similar learning curves for all group sizes, demonstrating that communication is successful even in larger groups. 
In all cases, interactive and self-communication test accuracy start low (25\%), but agents collaborate and end up between 60\% and 80\% success at $\mathsf{inter\_turn}$ = 100. 


For production preferences, we plot proportion of marking by \textit{order entropy} as we are again interested in order flexibility rather than the specific order chosen by the agents (\Cref{fig:group-results}, right column). Here, each circle denotes the average production preferences of an entire group, as opposed to those of a single agent. When comparing results across different group sizes, we see that the variability observed in self-playing agents (\Cref{sec:exp_selfComm}) including less optimal and redundant strategies, gets smaller as group size increases. The average entropy in groups of 8 and 20 is also lower than in groups of 4 or 2.  In the group setting, an agent's choice to use a marker does not only depend on its own order entropy but on that of the entire group. As a measure of the order/marking trade-off at group level, we therefore calculate Spearman's correlation ($\rho$) between order entropy and marker use, both computed over all (categorizable) utterances produced by all agents in a group. 
As shown in \Cref{fig:group-results}, $\rho$ steadily increases with group size from relatively weak (0.32) in pairs to strong (0.73) in groups of 20. This confirms that pairs, like self-playing agents, still often settle on redundant strategies, while larger groups develop more optimized languages in which stronger order consistency at the group level leads to a drop in marker use, confirming the emergence of the trade-off also at the group level.\footnote{Even when trained for much longer, the results of pairs remain similar, suggesting they indeed settle on less optimized solutions which is not overcome simply by more interactions (e.g. 200 rounds, $\rho$ = 0.33). See \Cref{app:extra-turns}.}

\section{Discussion and conclusion}
We introduced NeLLCom-X, a framework for simulating neural agent language learning and communication in groups, starting from pre-defined languages. Agents in this framework display the cognitively plausible property of interchangeability \cite{hockett1960origin}, by which anything they can understand, they can say and vice versa, while also having the ability to align to other individuals. We replicated an earlier finding by \citet{lian-etal-2023-communication} and showed that a word-order/case-marking trade-off still appears with the adjusted full-fledged agent architecture. 
Subsequently, we simulated interactions between agents trained on different languages. 
We found that pairs quickly adapt their utterances towards a mutually understandable language and that the neutral language drifts in different directions depending on the preferences of the other agent. Moreover, agents converge on a shared language faster, and reach higher accuracy in cases where one of the two agents has a stronger word order preference. 
We then assessed the effect of performing self-play during interactive communication and found it necessary to ensure our full-fledged agents continue to understand themselves, while also realistically adapting to other individuals. 
%
Lastly, we studied group dynamics and found that NeLLCom-X agents manage to establish a successful communication system even in larger groups (up to size 20).
Moreover, we generally see a larger entropy reduction in the languages developed by larger groups as compared to the languages used by pairs of agents. This finding aligns with previous work on group-level emergent communication, where it was shown that groups developed less idiosyncratic languages than pairs \cite{tieleman2019shaping} as well as with human experiments which demonstrated more systematic languages to emerge in larger groups \cite{raviv2019larger}. In our simulations, pairs and smaller groups sometimes settle on less optimized and partly still redundant solutions, while large groups end up with more efficient communication systems. 





In the future, NeLLCom-X can be used to study the influence of learning and group dynamics on many other language universals. We plan to keep refining the framework to allow studying different connectivities between the agents, multilingual populations and generational transmission of emerged languages to new agents. 

\section*{Limitations} 


Although the use of miniature artificial languages in our work allows for easily interpretable results due to abstractions and simplifications that are hard to achieve with natural human languages, the languages used currently are very small. This may limit the possibility of drawing conclusions beyond proof-of-concept demonstrations. Future work should increase the size and complexity of the languages to see if results hold on a larger scale and compare to patterns found in real human languages, such as those reported by \citet{levshina2023we}.


The meanings in our simulations are also strongly abstracted away from reality. While our design is well suited for an investigation of the word-order/case-marking trade-off, future simulations may need a less constrained meaning space, possibly using images to represent meanings.


All experiments conducted so far with NeLLCom-X use the same neural agent architecture (GRU), but we know that different architectures exhibit different inductive biases \cite{kuribayashi2024emergent} or memory constraints and these factors may influence the findings. 
Different types of neural learners, however, can be easily plugged into NeLLCom-X.


Interaction between individuals in groups is not the only population factor that shapes language, but linguistic structure is shaped by both interaction and learning \cite{kirby2015compression}. Especially when languages are learned and transmitted to subsequent generations repeatedly, even small inductive biases may have a large effect on emerging properties \cite{thompson2016culture}. We therefore plan to augment NeLLCom-X with iterated learning so that new agents learn from the utterances of others and become teachers to agents in the next generation.  

Finally, our agents are interacting in groups with multiple individuals, but they currently do not have any awareness of agent identities. A more realistic simulation should take into account that individuals know who they are interacting with, which becomes even more important when different network structures and connectivities will be explored.



\section*{Acknowledgements}
Arianna Bisazza acknowledges the support of the Dutch Research Council (NWO) within the InDeep project (NWA.1292.19.399) and the Talent Programme (VI.Vidi.221C.009).

\bibliography{acl_latex}

\newpage

\appendix

\section{More Details about NeLLCom}
\label{app:nellcom-details}

We list here additional details on the original NeLLCom framework \cite{lian-etal-2023-communication} that also apply to our extended NeLLCom-X framework.

\paragraph{Speaker and Listener Architectures}
Both speaking and listening networks have a single 16-dim GRU layer. The shared meaning embeddings have 8-dim and the shared word embeddings have 16-dim. The maximum utterance length for the speaking decoder is set to 10 words.

\paragraph{Supervised Language Learning}
During supervised learning, the speaker learns the mapping from the meaning inputs to utterances and vice versa for the listener.
Dataset $D$ is composed of meaning-utterance pairs $(m,u)$ where $u$ is the gold-standard generated for $m$ by a predefined grammar.
Given training sample $(m,u)$, speaker's parameters $\theta_\mathcal{S}$ and listener's parameters $\theta_\mathcal{L}$ are optimized by minimizing the cross-entropy loss of the predicted words and the predicted meaning tuples respectively:
\begin{multline}
Loss^{sup}_{(\mathcal{S})}  = -\sum_{i=1}^{I} \log p_{\theta_\mathcal{S}}(w^i|w^{<i},m)
\end{multline}
\noindent 
\begin{multline}
Loss^{sup}_{(\mathcal{L})} = 
- (\log p_{\theta_\mathcal{L}}(A | u) \\
+ 
\log p_{\theta_\mathcal{L}}(a | u) + \log p_{\theta_\mathcal{L}}(p | u))
\label{fuc:lst_loss}
\end{multline}
\noindent 
where $w_i ...  w_I$ are the words composing utterance $u$, whereas
$A,a,p$ are respectively the action, agent and patient of meaning $m$.

\paragraph{Communicative Reward Optimization}
Communication is implemented by a meaning reconstruction game following common practice in the artificial agent communication literature \citep[e.g.][]{steels1997synthetic,lazaridou2018emergence}.
The speaker generates an utterance $\hat{u}$ given a meaning $m$, and the listener needs to reconstruct meaning $m$ given $\hat{u}$.
The policy-based algorithm REINFORCE \cite{williams1992simple} is used to maximize a shared reward 
$r^\mathcal{L}(m, \hat{u})$, defined as the log likelihood of $m$ given $\hat{u}$ according to the listener's model: 
\begin{equation}
r^\mathcal{L}(m, \hat{u}) = \sum_{e\ \in\ m=\{A, a, p\}} \log p_{\theta_\mathcal{L}}(e | \hat{u})
\label{fuc:comm_r}
\end{equation}
Thus, the communication loss becomes:
\begin{equation}
    Loss_{(\mathcal{S,L})}^{comm} = -r^\mathcal{L}(m, \hat{u})*\sum_{i=1}^{I} \log p_{\theta_\mathcal{S}}(w^i|w^{<i},m)
\end{equation}

\begingroup
\setlength{\tabcolsep}{0pt} 
\renewcommand{\arraystretch}{0.6}
\begin{table*}[h!]
  \centering
  \begin{tabular}{c c c c c}

  & \small (1) Comm. success
  & \small (2) Order use
  & \small (3) Cond. Marker use
  & \small (4) Marker user
  \\
  \rotatebox[origin=c]{90}{\small (a) 100s+67m} &
  \begin{minipage}{.24\textwidth}
      \includegraphics[width=\columnwidth]{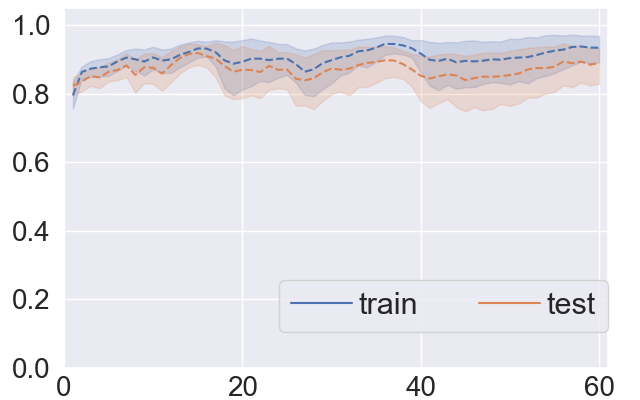}
    \end{minipage}
    &
    \begin{minipage}{.24\textwidth}
      \includegraphics[width=\columnwidth]{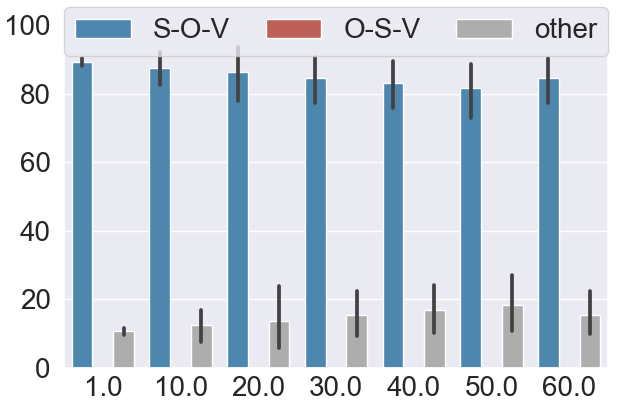}
    \end{minipage}
    &
    \begin{minipage}{.24\textwidth}
      \includegraphics[width=\columnwidth]{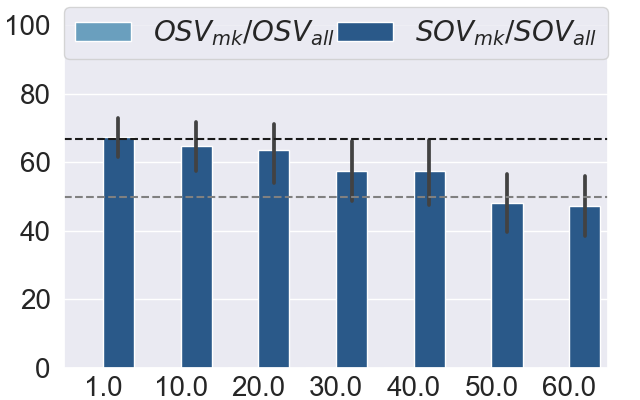}
    \end{minipage}
    &
    \begin{minipage}{.24\textwidth}
      \includegraphics[width=\columnwidth]{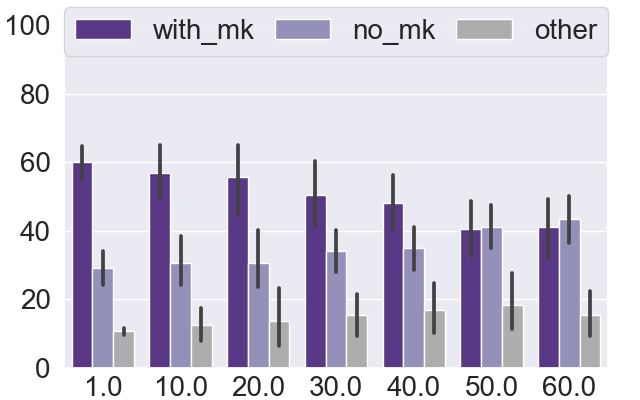}
    \end{minipage}
    \\

    \rotatebox[origin=c]{90}{\small (b) 100s+50m} &
  \begin{minipage}{.24\textwidth}
      \includegraphics[width=\columnwidth]{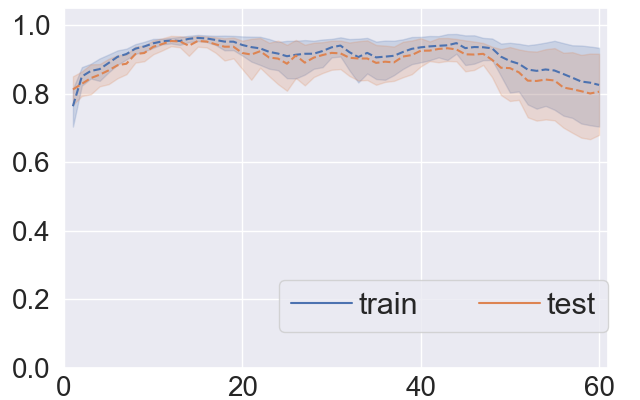}
    \end{minipage}
    &
    \begin{minipage}{.24\textwidth}
      \includegraphics[width=\columnwidth]{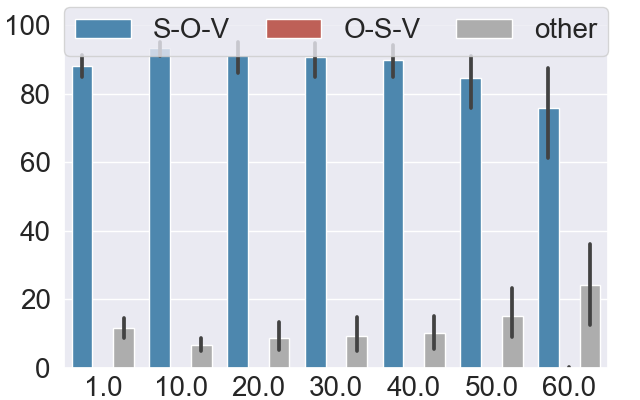}
    \end{minipage}
    &
    \begin{minipage}{.24\textwidth}
      \includegraphics[width=\columnwidth]{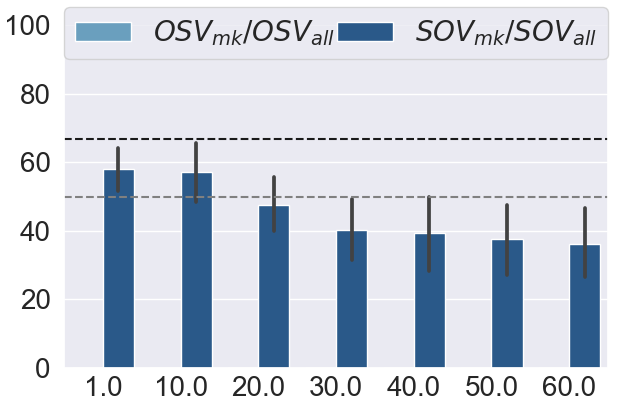}
    \end{minipage}
    &
    \begin{minipage}{.24\textwidth}
      \includegraphics[width=\columnwidth]{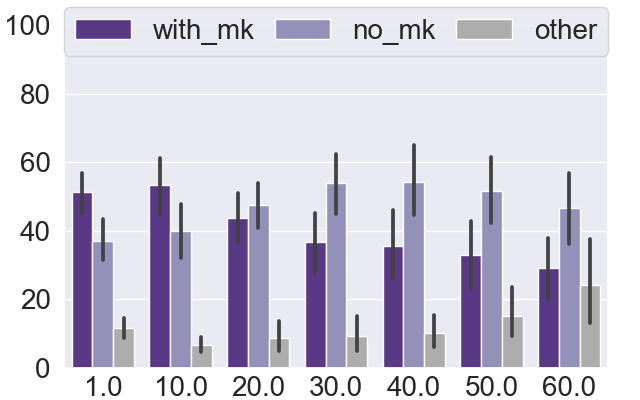}
    \end{minipage}
    \\

\rotatebox[origin=c]{90}{\small (c) 50s+67m} &
  \begin{minipage}{.24\textwidth}
      \includegraphics[width=\columnwidth]{fig/fig_results/free-op_group1_mk67osv50_Accuracy_Self-comm_phase0.png}
    \end{minipage}
    &
    \begin{minipage}{.24\textwidth}
      \includegraphics[width=\columnwidth]{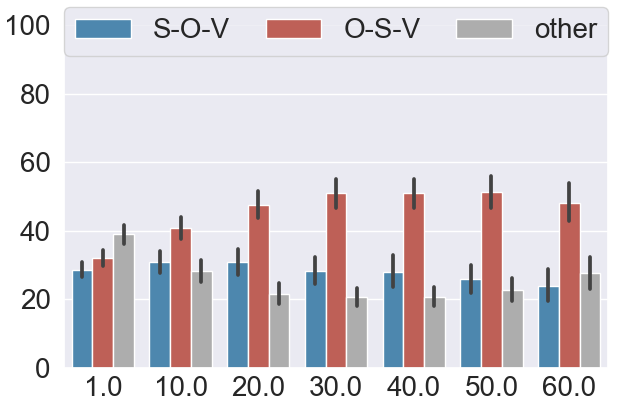}
    \end{minipage}
    &
    \begin{minipage}{.24\textwidth}
      \includegraphics[width=\columnwidth]{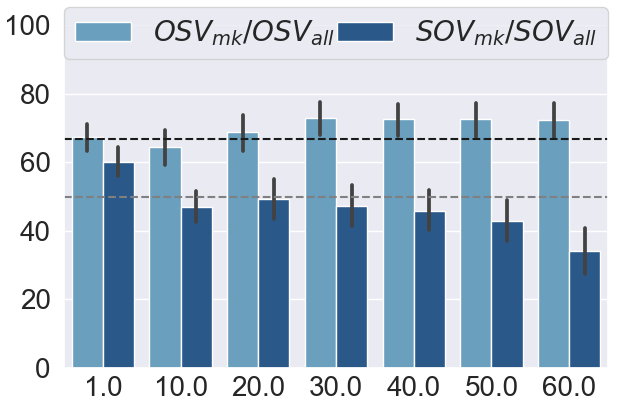}
    \end{minipage}
    &
    \begin{minipage}{.24\textwidth}
      \includegraphics[width=\columnwidth]{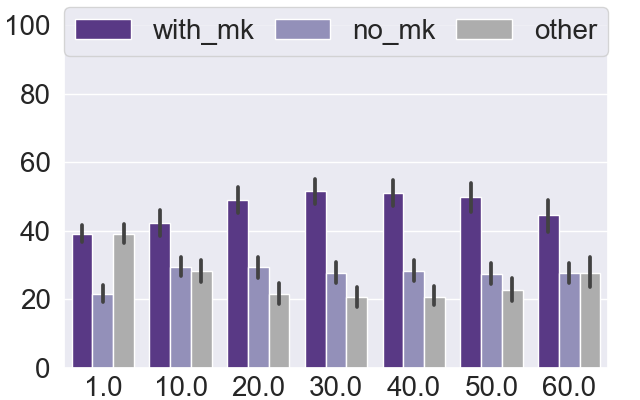}
    \end{minipage}
    \\

\rotatebox[origin=c]{90}{\small (d) 50s+50m} &
  \begin{minipage}{.24\textwidth}
      \includegraphics[width=\columnwidth]{fig/fig_results/free-op_group1_mk50osv50_Accuracy_Self-comm_phase0.png}
    \end{minipage}
    &
    \begin{minipage}{.24\textwidth}
      \includegraphics[width=\columnwidth]{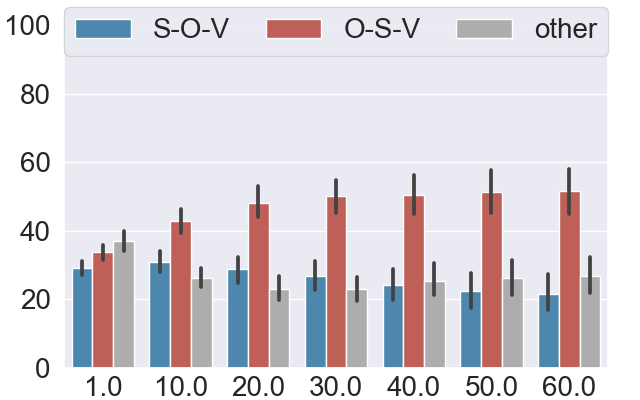}
    \end{minipage}
    &
    \begin{minipage}{.24\textwidth}
      \includegraphics[width=\columnwidth]{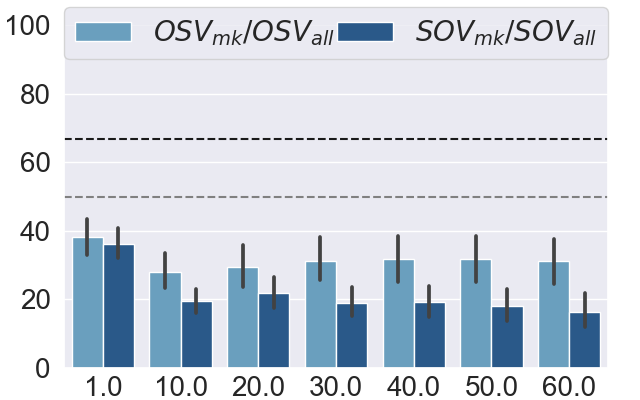}
    \end{minipage}
    &
    \begin{minipage}{.24\textwidth}
      \includegraphics[width=\columnwidth]{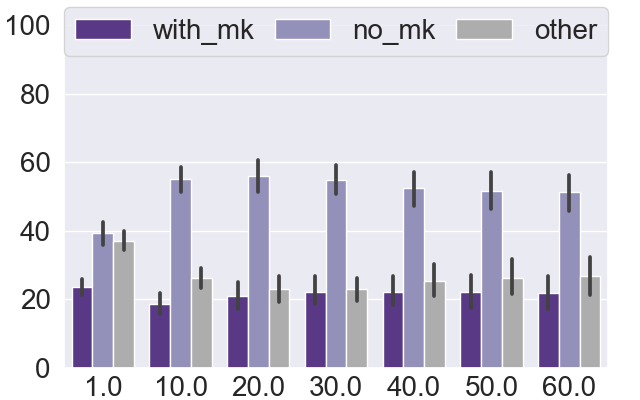}
    \end{minipage}

  \end{tabular}
  \captionof{figure}
  {
  Replicating the results from \citet{lian-etal-2023-communication} with NeLLCom-X full-fledged self-communicating agents with fixed-order (a) and flexible order (c) languages. Comparing the original results with a new, more neutral, initial languages with 50\% markers in (b) and (d). 
}
\caption*{}
\vspace{-2em}
\label{tab:replication}
\end{table*}
\endgroup

\begin{figure*}[h]
\centering
\captionsetup{singlelinecheck=off}
\begin{subfigure}{0.34\textwidth}
  \centering
  \includegraphics[width=\columnwidth]{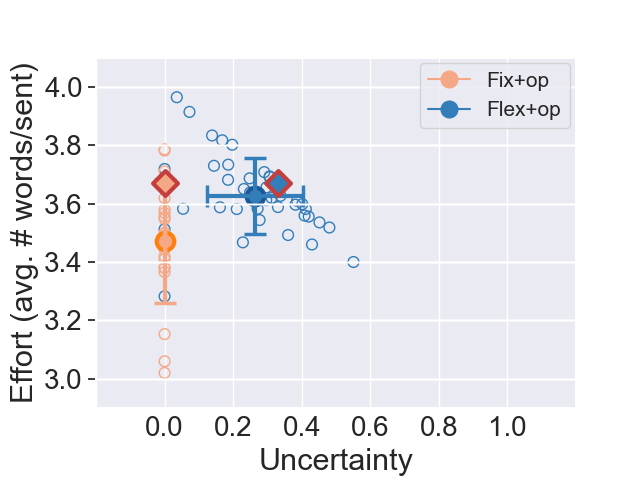}
  \caption{NeLLCom-X self-communication}
  \label{fig:diamond_selfcomm}
\end{subfigure}%
\begin{subfigure}{0.34\textwidth}
  \centering
  \includegraphics[width=\columnwidth]{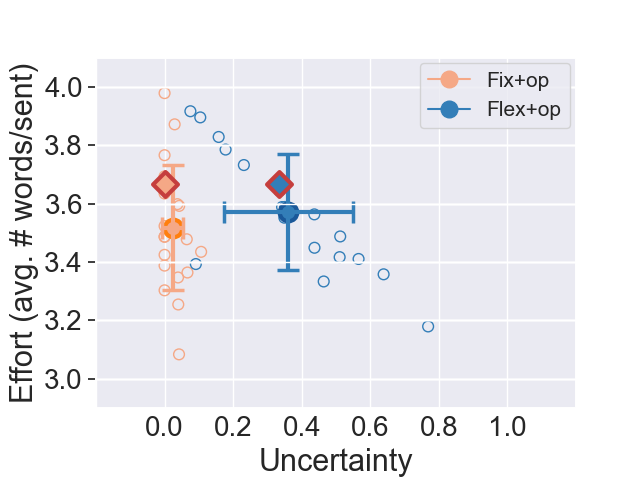}
  \caption{\citet{lian-etal-2023-communication} communication}
  \label{fig:diamond_tacl}
\end{subfigure}%
\begin{subfigure}{0.3\textwidth}
  \centering
  \includegraphics[width=\columnwidth]{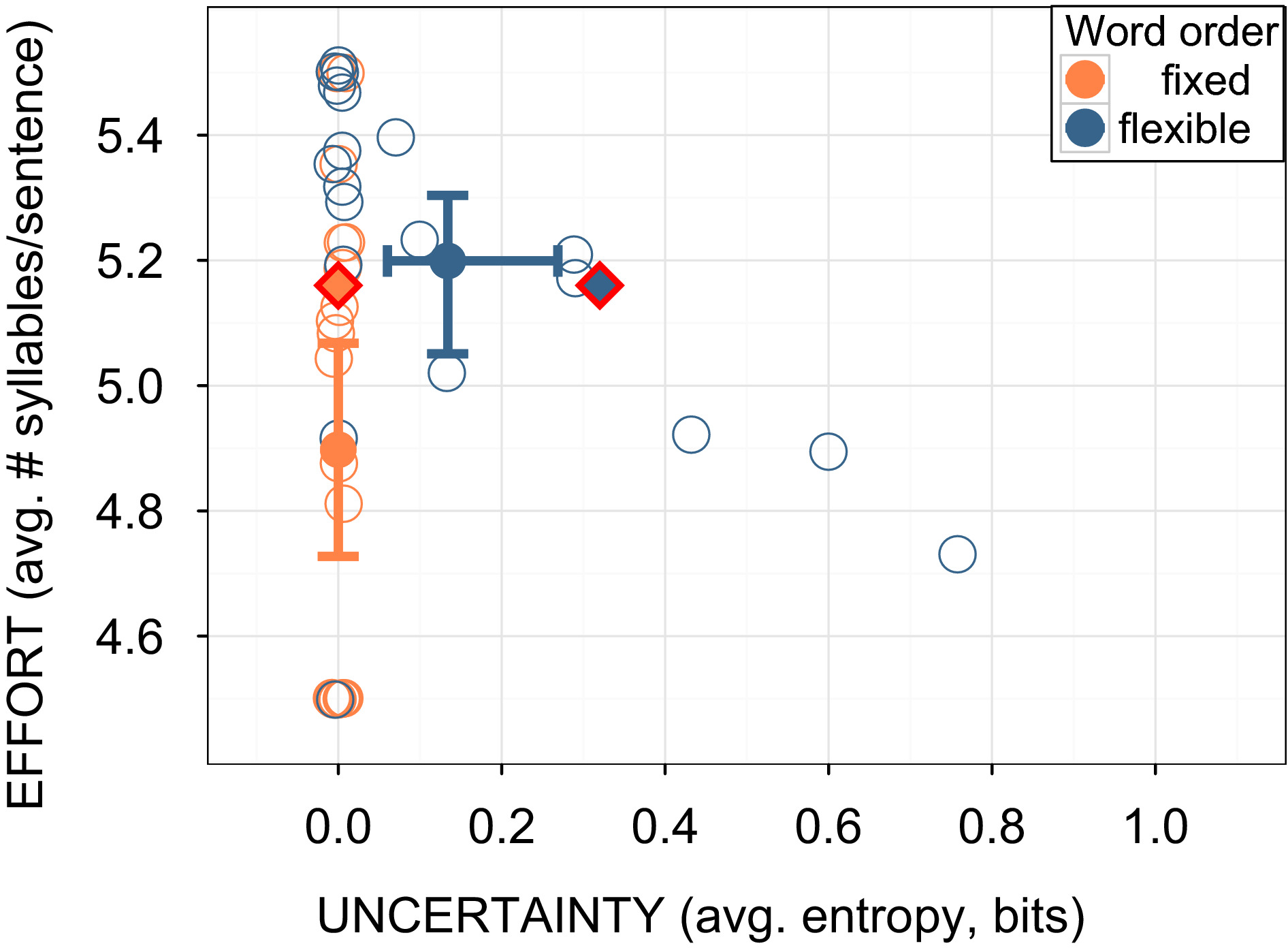}
  \caption{Humans \cite{fedzechkina2017balancing}}
  \label{fig:diamond_human}
\end{subfigure}
\begin{subfigure}{\textwidth}
  \centering
  \includegraphics[width=\textwidth, height=5cm]{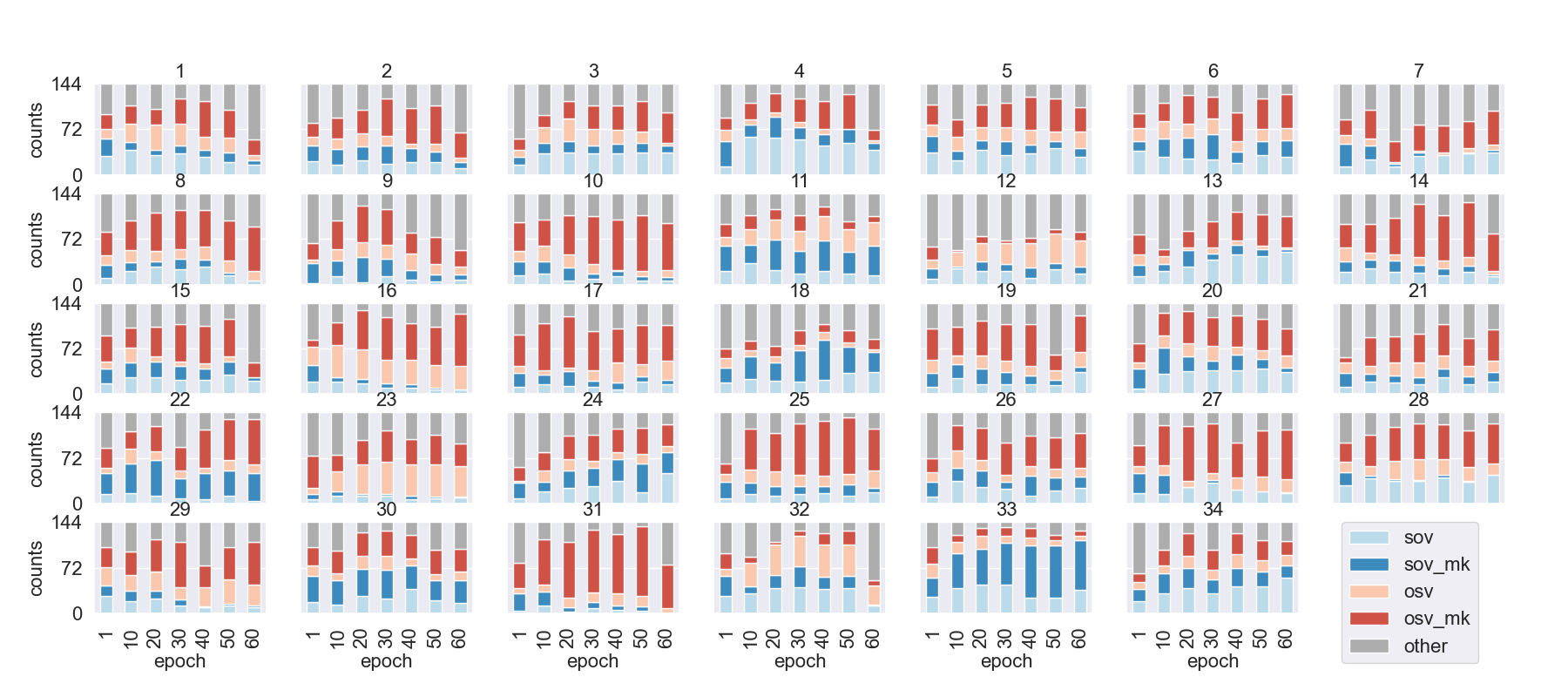}
  \caption{flex-mk67: Individual production patterns during self-communication.
  }
  \label{fig:self_train_dist}
\end{subfigure}
\caption{
Replicating the results of \citet{lian-etal-2023-communication}: Supervised learning followed by Self-communication with NeLLCom-X full-fledged agents.
All results are averaged over 50 random seeds.
}
\label{fig:replicate_full}
\end{figure*}

\section{Replicating NeLLCom Results with NeLLCom-X Full-fledged Agents}
\label{app:replicateTacl}

\subsection{Training details for the replication}
\label{app:replicat_traindetail}
For this replication (discussed in \Cref{sec:exp_selfComm}), we make the training configuration as consistent as possible with \citet{lian-etal-2023-communication}. Specifically, we split the data into 66.7/20\% training/testing. The testing proportion is different from the 33.3\% used in NeLLCom as we would like to match the test set size we use for interactive communication in this work. All entities and actions are required to appear at least once in the training set.
The default Adam optimizer is applied with a learning rate of 0.01. Both SL and $\mathsf{self\_turn}$ iterate 60 times.\footnote{As the 66.7\% trainset results in 480 samples, which equals 15 batches of 32 samples per turn. This is slightly different than 10 batches per turn during interactive communication.} Each replication setup is repeated with 50 random seeds.


\subsection{Results}

\paragraph{Fixed-order self-communication}
Starting from the initial marker proportion (66.7\%), fixed-order language learners start to drop the marker (50\% at round 60) during self-communication while maintaining high understandability (95\%) (\Cref{tab:replication} (a1) and (a4)). This aligns with the results of \citet{lian-etal-2023-communication}.

\paragraph{Flexible-order self-communication}

The self-communication accuracy in the flexible-order language (\Cref{tab:replication} (c1)) starts from a relatively low success rate as expected, but increases with more communication rounds. 
In particular, agents exceed the communication success they had achieved at the end of SL on new meanings and finally reach a much higher accuracy on new meanings at the end of self-communication (around 75\%) comparing to the communication success they had achieved at the end of SL.

The average ordering and marking proportions also show that flexible-order language self-communication results in a very similar pattern as was found by \citet{lian-etal-2023-communication}: 
(i) The average word order production (\Cref{tab:replication} (c2)) shows a strong preference for OSV, (ii) Although the overall marking system ends with a similar marker proportion as the initial condition (\Cref{tab:replication} (c4)), i.e., the proportion of with-marker utterances is twice the proportion of no-marker utterances, 
we can see a clear shift to conditional marking (\Cref{tab:replication} (c3)) with an asymmetric use of markers: at round 60, the marker proportion on utterances with OSV order (70\%) remains similar to the initial proportion (66.7\%), while the proportion of markers use with SOV drops to 35\%. This order preference and asymmetric marking system align with the flexible-order language results of \citet{lian-etal-2023-communication}.

\Cref{fig:self_train_dist} shows the production preferences of individual agents where the distributions of utterance type usage diverge over time, similar to the independent speaker and listener communication results in \citet{lian-etal-2023-communication}.


\paragraph{Uncertainty vs. Effort} 

\citet{lian-etal-2023-communication} found that agents balanced uncertainty and effort in a similar way to human participants in an artificial language learning task \cite{fedzechkina2017balancing}. To evaluate whether a similar uncertainty-effort trade-off is found with our full-fledged agents, we apply the same measurement on both fixed and flexible languages in \Cref{fig:diamond_selfcomm}. Besides the results from our new framework, we also reproduce the independent listener-speaker communication result from \citet{lian-etal-2023-communication} (\Cref{fig:diamond_tacl}) and human results from \citet{fedzechkina2017balancing} (\Cref{fig:diamond_human}) for comparison.

For the fixed-order language, the obvious drop of the averaged effort fits both \citet{lian-etal-2023-communication} and \citet{fedzechkina2017balancing}. 
Among 50 agents, only one agent significantly increases the use of markers and ends at around 3.8 words per utterance. Others reduce the marker, and two agents even end with 3.0 and 3.05 words per utterance which means almost no markers are produced.
For the flexible-order language, uncertainty is reduced slightly less strongly as in the human results, which was also the case in \cite{lian-etal-2023-communication}. 


\begin{table*}[htb]
\begin{center} \small
\begin{tabular}{c|l|l|l}
\hline \bf group size & \bf \# comm\_edges & \bf \# comm\_rounds & \bf \# repeated groups \\ 
\hline
2  & $2 = 2 * (2-1)$ & $100 = \ceil*{100 / (2-1)}$ & $100 = 200/2$ \\
\hline 4  & $12 = 4 * (4-1)$ & $34 = \ceil*{100 / (4-1)}$ & $50 = 200/4$ \\
\hline 8  & $56 = 8 * (8-1)$ & $15 = \ceil*{100 / (8-1)}$ & $25 = 200/8$ \\
\hline 20  & $380 = 20 * (20-1)$ & $6 = \ceil*{100 / (20-1)}$ & $10 = 200/20$ \\
\hline
\end{tabular}
\end{center}
\caption{Number of communication edges, number of rounds, and number of repeated groups for each group-size setting. Theaw settings were selected to ensure a fair comparison (i.e. similar amount of computation) across different group sizes.
\label{tab:diff_group}}
\end{table*}

\paragraph{50\% marking in initial language} 
As described in \Cref{par:initial_marking}, the initial proportion of marker use of 67\%, which was used in \citet{lian-etal-2023-communication} and inherited from \citet{fedzechkina2017balancing}, may create a bias for the agents to regularize towards more marker use, settling on more redundant languages. We therefore switched to the more neutral value of 50\% markers in the initial language. In \Cref{tab:replication}, the self-communication results of this new setting can be directly compared to the original set-up. As expected, markers are dropped more rapidly in the fixed-order 50\% marker language than in the 67\% marker language (\Cref{tab:replication} (a3) versus \Cref{tab:replication} (b3)). In the flexible-order languages, agents trained on the 67\% marker language mostly kept using the marker, even though they also developed a clear preference for one word order, resulting in redundant strategies. With 50\% markers in the initial language, however, agents drop the marker when they develop a word order preference despite being trained on a flexible word order language (\Cref{tab:replication} (c3) versus \Cref{tab:replication} (d3)).

\section{Training Details for Interactive Communication Experiments}\label{app:inter_traindetail}

We explain here the detailed setup for the main experiments discussed in \Cref{sec:exp_diffLangs} and \Cref{sec:exp_groupSize}.
This setup was determined based on preliminary experiments to yield optimal results in terms of learning accuracy (during SL) and communication success (during RL).

\paragraph{Data splits}
We first split the data into 80/20\% training/test. The test split is used thoughout the whole training. We resample 66.7\% meanings out of the first train set (resulting in 480 meaning-utterance pairs) for the SL training phase.
All entities and actions are required to appear at least once in the training set.

Then, for each communication turn, 50\% meanings are sampled from the first train set (resulting in 320 meanings) and used as the training samples for this RL turn.
%
Because interactive communication is always preceded by SL, agents have already learnt the mapping between words and entities and actions in the meaning space. Thus we do not enforce the all-seen-entities/actions rule in RL sampling.

\paragraph{Communication turns and rounds}

During interactive communication, the RL learning rate is set to 0.005.
For each communication turn, 1 epoch is applied corresponding to 10 batches of 32 meanings.
We fix the total number of inter\_turn per agent to (approximately) 200 (both speaking and listening are considered).
The total round is then computed as: 
$$comm\_rounds = \ceil*{\frac{200 * group\_size}{2 *|commu\_edges|}},$$ 
or to simplify the equation in fully connected communication graphs: 
$$comm\_rounds = \ceil*{\frac{100}{group\_size - 1}}.$$  
For a group of 2, a communication round includes 2 communication edges to be sampled: $\mathcal{G}_{g2} = \{\alpha_0 \to \alpha_1, \alpha_1 \to \alpha_0 \}$. 
For a group of 4, a communication round includes $12 = 4 \times (4-1)$ communication edges $\mathcal{G}_{g4} = \{A_0 \to A_1, A_0 \to A_2, A_0 \to A_3, A_1 \to A_0, A_1 \to A_2, A_1 \to A_3, A_2 \to A_0, A_2 \to A_1, A_2 \to A_3, A_3 \to A_0, A_3 \to A_1, A_3 \to A_2 \}$. 
Similarly, $|\mathcal{G}_{g8}| =8 \times (8-1) =56$
and $|\mathcal{G}_{g20}| = 20 \times (20-1) = 380$.
%
As for self-play, each agent performs 200/$\sigma$ self-play turns in total during interaction, that is 200/10=20 in the standard case where $\sigma=10$.


\paragraph{Number of random seeds}
In \Cref{sec:exp_diffLangs} we repeat each language combination experiment with 50 pairs of agents (i.e. 100 random seeds).
In \Cref{sec:exp_groupSize}, we set the total number of trained agents to 200 in each setup, (i.e. number of groups = $200/group\_size$). 
The details of rounds and repeated groups are listed in \Cref{tab:diff_group}.


\begingroup
\setlength{\tabcolsep}{0pt} 
\renewcommand{\arraystretch}{0.6}
\begin{table}[h!]
  \centering
  \begin{tabular}{c c c}

  & \shortstack{\small Communicative success \\ \small per turn}
  &
  \shortstack{\small Marker use \\ \small by order entropy}\\

  \rotatebox[origin=c]{90}{\small 100 rounds} &
  \begin{minipage}{.24\textwidth}
      \includegraphics[width=\columnwidth]{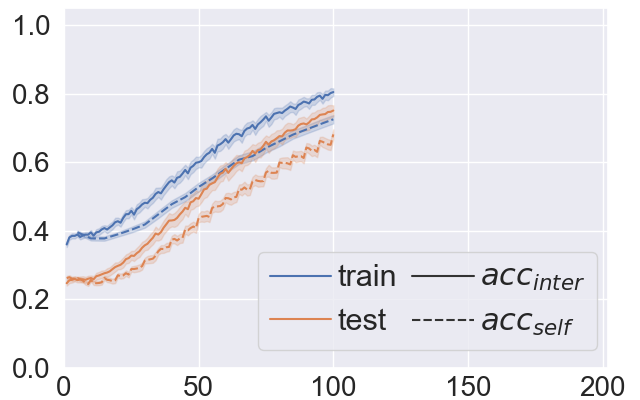}
    \end{minipage}
    &
    \begin{minipage}{.24\textwidth}
      \includegraphics[width=\columnwidth]{fig/fig_results/free-op_group2_mk50osv50_H_Order+mk_group.png}
    \end{minipage}
    \\

    \rotatebox[origin=c]{90}{\small 200 rounds} &
  \begin{minipage}{.24\textwidth}
      \includegraphics[width=\columnwidth]{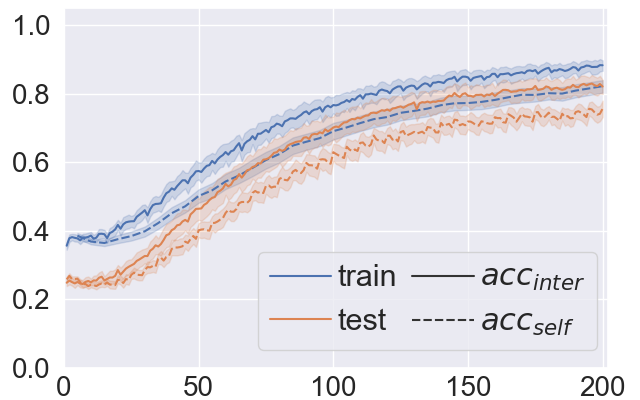}
    \end{minipage}
    &
    \begin{minipage}{.24\textwidth}
      \includegraphics[width=\columnwidth]{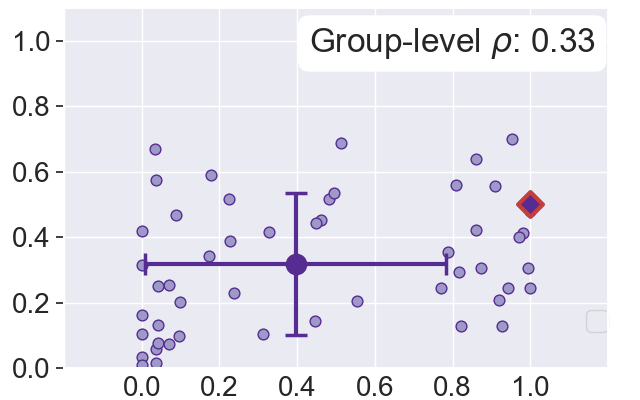}
    \end{minipage}

  \end{tabular}
  \captionof{figure}
  {
Interactive communication in pairs of same-language speakers (50s+50m): Production preferences (right column) do not change much when training for 200 rounds (bottom row) instead of 100 (top).
}
\caption*{}
\vspace{-2em}
\label{fig:group-results-extra-turns}
\end{table}
\endgroup

\section{Additional Group Experimennts}
\label{app:extra-turns}

\Cref{fig:group-results-extra-turns} shows the effect of longer training on the production preferences of pairs of same-language speakers (50s+50m). 
Production preferences (right column) do not change much after 100 additional turns (bottom row), and the correlation $\rho$ increases only marginally from 0.32 to 0.33.


\end{document}